\documentclass{article}

\usepackage{times}

\usepackage{graphicx} 
\usepackage{subfigure} 

\usepackage{tikz}
\usetikzlibrary{shapes.geometric}
\usepackage{pgfplots,pgfplotstable}
\usepackage{natbib}

\usepackage{algorithm}
\usepackage{algorithmic}

\usepackage{amsmath}

\usepackage[accepted]{icml2014}

\usepackage{enumerate}
\usepackage{paralist}
\usepackage{color}
\usepackage{listings}
\lstdefinelanguage{scala}{
  morekeywords={abstract,bayes,case,catch,class,def,%
    do,else,extends,false,final,finally,%
    for,if,implicit,import,match,mixin,%
    new,null,object,override,package,%
    private,protected,requires,return,sealed,%
    super,this,throw,trait,true,try,%
    type,val,var,while,with,yield},
  otherkeywords={=>,<-,<\%,<:,>:,\#,@},
  sensitive=true,
  morecomment=[l]{//},
  morecomment=[n]{/*}{*/},
  morestring=[b]",
  morestring=[b]',
  morestring=[b]"""
}
\newcommand\eg{e.g.}

\icmltitlerunning{Augur: a Modeling Language for Data-Parallel Probabilistic Inference}

\begin{document} 

\twocolumn[
\icmltitle{Augur: a Modeling Language for Data-Parallel Probabilistic Inference}

\icmlauthor{Jean-Baptiste Tristan}{jean.baptiste.tristan@oracle.com}
\icmladdress{Oracle Labs}
\icmlauthor{Daniel Huang}{dehuang@fas.harvard.edu}
\icmladdress{Harvard University}
\icmlauthor{Joseph Tassarotti}{jtassaro@cs.cmu.edu}
\icmladdress{Carnegie Mellon University}
\icmlauthor{Adam Pocock}{adam.pocock@oracle.com}
\icmladdress{Oracle Labs}
\icmlauthor{Stephen J. Green}{stephen.x.green@oracle.com}
\icmladdress{Oracle Labs}
\icmlauthor{Guy L. Steele, Jr}{guy.steele@oracle.com}
\icmladdress{Oracle Labs}

\icmlkeywords{Probabilistic Programming, Markov Chain Monte Carlo, Bayesian Networks, Data-Parallel Algorithms, GPU}

\vskip 0.3in
]

\begin{abstract} 
It is time-consuming and error-prone to implement inference procedures
for each new probabilistic model.  Probabilistic programming addresses
this problem by allowing a user to specify the model and having a
compiler automatically generate an inference procedure for it. For
this approach to be practical, it is important to generate inference
code that has reasonable performance. In this paper, we present a
probabilistic programming language and compiler for Bayesian networks
designed to make effective use of data-parallel architectures such as
GPUs. We show that the compiler can generate data-parallel inference
code scalable to thousands of GPU cores by making use of the
conditional independence relationships in the Bayesian network. 
\end{abstract} 

\section{Introduction}
\label{Introduction}

Machine learning, and especially probabilistic modeling, can be
difficult to apply.  A user needs to not only design the model, but
also implement the right inference procedure.  There are many
different inference algorithms, most of which are conceptually
complicated and difficult to implement at scale. Despite the
enthusiasm that many people who practice data analysis have for
machine learning, this complexity is a barrier to deployment.  Any
effort to simplify the use of machine learning would thus be very
useful.

Probabilistic programming \cite{General}, as introduced in the BUGS
project \cite{BUGSOriginal}, is a way to simplify the application of
machine learning based on Bayesian inference. The key feature of
probabilistic programming is separation of concerns: the user
specifies {\it what} needs to be learned by describing a probabilistic
model, while the compiler automatically generates the {\it how}, that
is, the inference procedure. In particular, the programmer writes code
that describes a probability distribution. Using a compiler-generated
inference algorithm, the programmer then samples from this
distribution.

However, doing inference on probabilistic programs is computationally
intensive and challenging. As a result, developing algorithms to
perform inference is an active area of research. These include
deterministic approximations (such as variational methods) and Monte
Carlo approximations (such as MCMC algorithms).  The problem is that
most of these algorithms are conceptually complicated, and it is not
clear, especially for non-experts, which one would work best for a
given model.

To address the performance issues, our work has been driven by two
observations. The first observation is that good performance starts
with an appropriate inference algorithm, and selecting the right
algorithm is often the hardest problem. For example, if our compiler
only emits Metropolis-Hastings inference, there are models for which
our programming language will be of no use, even given large amounts
of computational power. We must design the compiler in such a way that
we can include the latest research on inference while reusing
pre-existing analyses and optimizations, or even mix inference
techniques. Consequently, we have designed our compiler as a modular
framework where one can add a new inference algorithm while reusing
already implemented analyses and optimizations. For that purpose, our
compiler uses an intermediate representation (IR) for probability
distributions that serves as a target for modeling languages and as a
basis for inference algorithms. We will show this IR is key to scaling
the compiler and the inference to very large networks.


The second observation is if we wish to continue to benefit from
advances in hardware we \emph{must} focus on producing highly
\emph{parallel} inference algorithms.  We claim that many MCMC
inference algorithms are highly data-parallel \cite{Steele,WorkSpan}
within a single Markov Chain if we take advantage of the
conditional independence relationships of the input model (e.g. the
assumption of i.i.d. data makes the likelihood independent across data
points).  Moreover, we can automatically generate good data-parallel
inference with a compiler.  Such inference will run very efficiently
on highly parallel architectures such as Graphics Processing Units
(GPUs). It is important to note that parallelism brings an interesting
trade-off for performance since some inference techniques can result
in less parallelism and will not scale as well.

In this paper, we present our compilation framework, named {\it
  Augur}. To start (Section \ref{Example}), we demonstrate how to
specify two popular models in our language and use them to do learning
and prediction. Then, (Section \ref{Design}), we describe how we
support different modeling languages by embedding them into Scala
using Scala's macro system, which provides type checking and IDE
support, and we describe the probabilistic IR.  Next, (Section
\ref{Generation}), we describe data-parallel versions of
Metropolis-Hastings and Gibbs sampling that scale on the GPU and speed
up sampling. Our compiler also includes a Metropolis-Within-Gibbs
sampler but we do not detail these in this paper. Then (Section
\ref{Experiments}), we present the results of some benchmarks, which
include comparisons against other implementations of inference for a
regression, a Gaussian Mixture Model, and LDA. Finally (Section
\ref{Related}), we review the literature in probabilistic programming.

Our main results are: first, not only are some inference algorithms
highly data-parallel and amenable to GPU execution, but a compiler can
{\it automatically} generate such GPU implementations effectively;
second, for the compiler to be able to handle large model
specifications (such as LDA) it is key to use a symbolic
representation of the distribution rather than constructing the
graphical model.

\section{The Augur Language}
\label{Example}

As examples, we first present the specification of two models in
Augur, Latent Dirichlet Allocation (LDA) \cite{LDA} and multivariate
regression. Then we show how the LDA model is used to learn the topics
present in a set of documents. The supplementary material contains
five examples of probabilistic models in Augur including a polynomial
regression, a categorical and a Gaussian Mixture Model (GMM), a Naive Bayes
classifier, and a Hidden Markov Model (HMM).

\subsection{Specifying the Models}

The LDA model specification is shown in Figure \ref{model}. The
probability distribution is defined as a Scala object ({\tt object
  LDA}) and is composed of two declarations. First, we declare the
support of the probability distribution as a class that must be named
{\tt sig}. Here the support is composed of four arrays, with one each
for the distribution of topics per document ({\tt theta}), the
distribution of words per topic ({\tt phi}), the topics assigned to
the words ({\tt z}), and the words in the corpus ({\tt w}). The
support is used to store the inferred model parameters. These last two
arrays are flat representations of ragged arrays, and so we do not
require the documents to be of equal length.

\begin{figure}[t]
\vskip 0.2in
\begin{center}
%
%
%
%
%
%
%
\begin{minipage}{0.93\linewidth}
\begin{lstlisting}
object LDA {
class sig(var phi : Array[Double], 
             var theta : Array[Double], 
             var z : Array[Int], 
             var w : Array[Int])

val model = bayes {
 (K: Int, V: Int, M: Int, N: Array[Int]) => {
   val alpha = vector(K,0.1)
   val beta = vector(V,0.1)         
   val phi = Dirichlet(V,beta).sample(K)   
   val theta = Dirichlet(K,alpha).sample(M) 
   val w = 
     for(i <- 1 to M) yield {
       for(j <- 1 to N(i)) yield {
         val z: Int = 
           Categorical(K,theta(i)).sample()
         Categorical(V,phi(z)).sample()
        }}
   observe(w)
}}}
\end{lstlisting}
\end{minipage}
\caption{Specification of the latent Dirichlet allocation model in Augur. The model specifies the probability distribution
  $p(\phi,\theta,z \mid w)$. The keyword {\tt bayes} introduces the modeling language for Bayesian networks. }
\label{model}
\end{center}
\vskip -0.2in
\end{figure}

The second declaration specifies the Bayesian network associated with
LDA and makes use of our domain specific language for Bayesian
networks. The DSL is marked by the {\tt bayes} keyword and delimited
by the following enclosing brackets. The model first declares the
parameters of the model: {\tt K} for the number of topics, {\tt V} for
the vocabulary size, {\tt M} for the number of documents, and {\tt N}
for the array that associates each document with its size.

In the model itself, we define the hyper-parameters (values {\tt
  alpha} and {\tt beta}) for the Dirichlet distributions and draw {\tt
  K} Dirichlet samples of dimension {\tt V} for the distribution of
words per topic ({\tt phi}) and {\tt M} Dirichlet samples of dimension
{\tt K} for the distribution of topics per document ({\tt
  theta}). Then, for each word in each document, we draw a topic {\tt
  z} from {\tt theta}, and finally a word from {\tt phi} based on the
topic we drew for {\tt z}.
 
\begin{figure}[t]
\vskip 0.2in
\begin{center}
\begin{minipage}{0.93\linewidth}
\begin{lstlisting}
object LinearRegression {
class sig(var w: Array[Double], var b: Double, 
             var tau: Double, var x: Array[Double], 
             var y: Array[Double])

val model = bayes {
 (K: Int, N: Int, l: Double, u: Double) => {

  val w = Gaussian(0, 10).sample(K)
  val b = Gaussian(0, 10).sample()
  val tau = InverseGamma(3.0, 1.0).sample()
  val x = for(i <- 1 to N) 
          yield Uniform(l, u).sample(K)
  val y = for (i <- 1 to N) yield {
      val phi = for(j <- 1 to K) yield w(j) * x(i)(j)
      Gaussian((phi.sum) + b, tau).sample()
    }        
  observe(x, y)
}}}
\end{lstlisting}
\end{minipage}
\caption{Specification of a multivariate regression in Augur.}
\label{regression}
\end{center}
\vskip -0.2in
\end{figure}

The regression model in Figure \ref{regression} is defined in the same
way and uses similar language features. In this example the support
comprises the ({\tt x},{\tt y}) data points, the weights {\tt w}, the
bias {\tt b}, and the noise {\tt tau}. The model uses an additional
{\tt sum} function to sum across the feature vector.

\subsection{Using the model}
\label{Evaluating}

Once a model is specified, it can be used as any other Scala object by
writing standard Scala code. For instance, one may want to use the LDA
model with a training corpus to learn a distribution of words per
topic and then use it to learn the per-document topic distribution of
a test corpus. An implementation is presented in Figure \ref{use}.
First the programmer must allocate the parameter arrays which contain
the inferred values. Then the signature of the model is constructed
which encapsulates the parameters. The {\tt LDA.model.map} command
returns the MAP estimate of the parameters given the observed words.

\begin{figure*}[t]
\begin{center}
\begin{minipage}{0.9\textwidth}
\begin{lstlisting}
val phi = new Array[Double](k * v)
val theta_train = new Array[Double](doc_num_train * k)
val z_train = new Array(num_tokens_train) 
val s_train = new LDA.sig(phi, theta_train, z_train, w_train)
LDA.model.map(Set(), (k, v, doc_num_train, docs_length_train), s_train, samples_num, Infer.GIBBS)
        
val z_test = new Array(num_tokens_test) 
val theta_test = new Array[Double](doc_num_test * k)
val s_test = new LDA.sig(phi, theta_test, z_test, w_test)
LDA.model.map(Set("phi"), (k, v, doc_num_test, docs_length_test), s_test, samples_num, Infer.GIBBS)
\end{lstlisting}
\end{minipage}
\caption{Example use of the LDA. Function {\tt LDA.model.map} returns
  a maximum a posteriori estimation. It takes as arguments the set of
  variables to observe (on top of the ones declared as observed in the
  model specification), the hyperparameters, the initial parameters,
  the output parameters, the number of iterations and the inference to
  use. The parameters are stored in {\tt LDA.sig}.}
\label{use}
\end{center}
\end{figure*}

To test the model, a new signature is constructed containing the test
documents, and the previously inferred {\tt phi} values. Then 
{\tt LDA.model.map} is called again, but with both the phis and the words
observed (by supplying {\tt Set("phi")}). The inferred thetas for 
the test documents are stored in {\tt s\_test.theta}.

\section{A Modular Compilation Framework}
\label{Design}

Before we detail the architecture of our compiler, it is useful to
understand how a model goes from a specification down to CUDA code
running on the GPU. There are two distinct compilation phases. The
first happens when the programmer compiles the program with scalac
(assuming that the code from Figure \ref{model} is in a file named
{\tt LDA.scala})

{\tt scalac -classpath augur.jar LDA.scala}

The file {\tt augur.jar} is the package containing our compiler. The
first phase of compilation happens statically, during normal {\tt
  scalac} compilation. In this phase, the block of code following the
{\tt bayes} keyword is transformed into our intermediate
representation for probability distributions.  The second compilation
phase happens at runtime, when the programmer calls the {\tt
  LDA.model.map} method. At that point, the IR is transformed,
analyzed, and optimized, and finally, CUDA code is emitted and run.

Our framework is therefore composed of two distinct components that
communicate through the IR: the front end, where domain specific
languages are converted into the IR, and the back end, where the IR
can be compiled down to various inference algorithms (currently
Metropolis-Hastings, Gibbs sampling, and Metropolis-Within-Gibbs). To
define a modeling language in the front end, we make use of the Scala
macro system. The macro system allows us to define a set of functions
(called ``macros'') that will be executed by the Scala compiler on the
code enclosed by the macro. We are currently focusing on Bayesian
networks, but other DSLs (\eg, Markov random fields) could be added
without modifications to the back end. The implementation of the
macros to define the Bayesian network language is conceptually
uninteresting so we omit further details. Our Bayesian
network language is fairly standard, with the notable exception that
it is {\it implicitly} parallel.

Separating the compilation into two distinct phases gives us many
advantages. As our language is implemented using Scala's macro system
it provides automatic syntax highlighting, method name completion and
code refactoring in any IDE which supports Scala. This greatly
improves the usability of the DSL as no special tools need to be
developed to support it. This macro system allows Augur to use Scala's
parser, semantic analyzer (\eg, to check that variables have
been defined), and type checker. Also we benefit from the Scala
compiler's optimizations such as constant folding and dead code
elimination.

Then, because the IR is compiled to CUDA code {\it at runtime}, we
know the values of all the hyper-parameters and the size of the
dataset.  This enables better optimization strategies, and also gives
us key insights into how to extract parallelism (Section
\ref{Gibbs}). For example, when compiling LDA, we know that the number
of topics is much smaller than the number of documents and thus
parallelizing over documents will produce more parallelism than
parallelizing over topics.

Finally, we also provide a library which defines standard
distributions such as Gaussian, Dirichlet, etc. In addition to these
standard distributions, each model denotes its own user-defined
distribution. All of these distributions are subtypes of the {\tt
  Dist} supertype. Currently, the {\tt Dist} interface provides two
methods: {\tt map}, which implements maximum a posteriori estimation,
and {\tt sample}, which returns a sequence of samples.


\section{Generation of Data-Parallel Inference}
\label{Generation}

%
%


When an inference procedure is invoked on a model (e.g. {\tt
  LDA.model.map}), the IR is compiled down to CUDA inference code for
that model. Informally, our IR expressions are generated from
this Backus-Naur form grammar:
\begin{gather*}
P ::= p(\stackrel{\rightarrow}{X}) \; \Big| \; p(\stackrel{\rightarrow}{X} |
\stackrel{\rightarrow}{X}) \; \Big| \; P P \; \Big| \; \frac{1}{P} \; \\ \Big| \; \prod\limits_i^N P \;
\Big| \; \int_X P \, {\it d} x \; \Big| \; \{ P \}_{c}
\end{gather*}
The goal of the IR is to make the sources of parallelism in the model
more explicit and to support analysis of the probability distributions
present in the model. For example, a $\prod$ indicates that each
sub-term can be evaluated in parallel.

The use of such a symbolic representation for the model is key to
scale to large networks. Indeed, as we will show in the experimental
evaluation (Section \ref{Experiments}), popular probabilistic
programming language implementations such as {\tt JAGS} or {\tt Stan}
reify the graphical model, resulting in unreasonable memory
consumption for models such as LDA.  A consequence of our symbolic
representation is that it becomes more difficult to discover conjugacy
relationships, a point we will come back to.

In the rest of this section, we explain how the compiler generates
data-parallel samplers that exploit the conditional independence
structure of the model. We will use our two examples to explain how
the compiler analyzes the model and generates the inference code.

\subsection{Generating data-parallel MH samplers}

If the user wants to use Metropolis-Hastings inference on a model,
the compiler needs to emit code for a function $f$ that is
proportional to the distribution the user wants to sample from. This
function is then linked with our library implementation of
Metropolis-Hastings. The function $f$ is composed of the product of
the prior and the likelihood of the model and is extracted
automatically from the model specification. For example, applied to
our regression example, $f$ is defined as
$$ f({\bf x},{\bf y},\tau,b,{\bf w}) = p(b) p(\tau) p({\bf w}) p({\bf
  x}) p({\bf y} | {\bf x}, b, \tau, {\bf w})
$$
which is equal to (and represented in our IR as)
$$ p(b) p(\tau) \left( \prod\limits_k^K p(w_k) \right) \left( \prod\limits_n^N p(x_n) p(y_n | {\bf x_n} \cdot {\bf w} + b, \tau \right)
$$
%
%
In this form, the compiler knows that the distribution factorizes into
a large number of terms that can be evaluated in parallel and then
efficiently multiplied together; more specifically, it knows that the
data is i.i.d. and that it can optimize accordingly. In this case,
each $(x,y)$ contributes to the likelihood independently, and they can
be evaluated in parallel. In practice, we work in log-space, so we
perform summations. The compiler can then generate the CUDA code for
the evaluation of $f$ from the IR representation. This code generation
step is conceptually simple and we will not explain it further.

It is interesting to note that despite the simplicity of this
parallelization the code scales reasonably well: there is a large
amount of parallelism because it is roughly proportional to the number
of data points; uncovering the parallelism in the code does not increase
the overall quantity of computation that has to be performed; and the
ratio of computation to global memory accesses is high enough to hide
memory latency bottlenecks.

\subsection{Generating data-parallel Gibbs samplers} \label{Gibbs}

Alternatively, and more interestingly, the compiler can generate a
Gibbs sampler for some models. For instance, we would like to generate
a Gibbs sampler for LDA, as a simple Metropolis-Hastings sampler will
have a very low acceptance ratio. Currently we cannot generate a
collapsed or blocked sampler, but there is interesting work related to
dynamically collapsing or blocking variables \cite{Dynamic}, and we
leave it to future work to extend our compiler with this capability.

To generate a Gibbs sampler, the compiler needs to Figure out how to
sample from each univariate distribution.  As an example, to draw
$\theta_m$ as part of the $(\tau+1)$th sample, the compiler needs to
generate code that samples from the following distribution
$$
p(\theta_m^{\tau + 1} | w^{\tau+1}, z^{\tau+1}, \theta_1^{\tau+1}, ..., \theta_{m-1}^{\tau + 1}, \theta_{m+1}^{\tau}, ..., \theta_M^{\tau})
$$
As we previously explained, our compiler uses a symbolic
representation of the model: the upside is that it makes it possible
to scale to large networks, but the downside is that it becomes more
challenging to uncover conjugacy relations and independence between
variables. To accomplish this, the compiler implements an algebraic
rewrite system that attempts to rewrite the above expression in terms
of expressions it knows (i.e., the joint distribution of the entire
model). We show a few selected rules below to give a flavor of the
rewrite system.
\begin{compactenum}[(a)]
\item $\frac{P}{P} \Rightarrow 1$ \\
\item $\int P(x) \, Q \, {\it d}x \Rightarrow Q \int P(x) dx$ \\
\item $\prod\limits_i^NP(x_i) \Rightarrow \prod\limits_i^N \{P(x_i)\}_{q(i) = true}
\prod\limits_i^N \{ P(x_i) \}_{q(i) = false}$ \\
\item $P(x) \Rightarrow \frac{P(x,y)}{\int P(x,y) \, {\it d}y}$
\end{compactenum}
Rule (a) states that like terms can be canceled. Rule (b) says that
terms that do not depend on the variable of integration can be pulled out
of the integral. Rule (c) says that we can partition a product over
N-terms into two products, one where a predicate $q$ is satisfied on
the indexing variable and one where it is not. Rule (d) is a
combination of the product and sum rule. Currently, the rewrite system
is just comprised of rules we found useful in practice, and it is easy
to extend the system to add more rewrite rules.

Going back to our example, the compiler rewrites the desired
expression into the one below: 
$$ \frac{p(\theta_m^{\tau+1}) \prod\limits_j^{N(m)} p(z_{mj} |
  \theta_m^{\tau+1})}{\int p(\theta_m^{\tau+1}) \prod\limits_j^{N(m)}
  p(z_{mj} | \theta_m^{\tau+1}) d \theta_m^{\tau+1}}
$$ 
In this form, it is clear that each $\theta_1, \dots, \theta_m$
is independent of the others after conditioning on the other random
variables. As a result, they may all be sampled in parallel.

At each step, the compiler can test for a conjugacy relation. In the
above form, the compiler recognizes that the $z_{mj}$ are drawn from a
categorical distribution and $\theta_m$ is drawn from a Dirichlet, and
can exploit the fact that these are conjugate distributions. The
posterior distribution for $\theta_m$ is:
$$
{\tt Dirichlet}(\alpha + c_m)
$$
where $c_m$ is a vector whose $k$th entry is the number of $z$
associated with document $m$ that were assigned topic
$k$. Importantly, the compiler now knows that the drawing of each $z$
must include a counting phase.


%
%
%
%
%
%

The case of the $\phi$ variables is more interesting. In this case, we want to
sample from
$$
p(\phi_k^{\tau+1} | w^{\tau+1}, z^{\tau+1}, \theta^{\tau+1}, \phi_1^{\tau+1}, ..., \phi_{k-1}^{\tau+1}, \phi_{k+1}^{\tau}, ..., \phi_K^{\tau})
$$
After the application of the rewrite system to this expression, the compiler
discovers that this is equal to
$$
\frac{p(\phi_k) \prod\limits_i^M \prod\limits_j^{N(i)} \{ p(w_i | \phi_{z_{ij}}) \}_{k
  = z_{ij}}}{\int p(\phi_k) \prod\limits_i^M \prod\limits_j^{N(i)} \{ p(w_i | \phi_{z_{ij}}) \}_{k
  = z_{ij}} d\phi_k}
$$
The key observation that the compiler takes advantage of to reach this
conclusion is the fact that the $z$ are distributed according to a
categorical distribution and are used to index into the $\phi$ array.
Therefore, they partition the set of words $w$ into $K$ disjoint sets
$w_1 \uplus ...  \uplus w_k$, one for each topic. More concretely, the
probability of words drawn from topic $k$ can be rewritten in
partitioned form using rule (c) as
$$
\prod\limits_i^M \prod\limits_j^{N(i)} \{ p(w_{ij} | \phi_{z_{ij}}) \}_{k =z_{ij}}
$$ 
This expresses the intuition that once a word's topic is fixed, the
word depends on only one of the $\phi_k$ distributions. In this form,
the compiler recognizes that it should draw from
$$
{\tt Dirichlet}(\beta + c_k)
$$
where $c_k$ is the count of words assigned to topic $k$.  In general,
the compiler detects patterns like the above when it notices that
samples drawn from categorical distributions are being used to index
into arrays.

Finally, the compiler turns to analyzing the $z_{ij}$. In this case,
it will again detect that they can be sampled in parallel but it will
not be able to detect a conjugacy relationship. It will then detect
that the $z_{ij}$ are drawn from discrete distributions, so that the
univariate distribution can be calculated exactly and sampled from. In
cases where the distributions are continuous, it can try to use
another approximate sampling method as a subroutine for drawing that
variable.

One concern with such a rewrite system is that it may fail to find a
conjugacy relation if the model has a complicated structure. So far we
have found our rewrite system to be robust and it can find all the
usual conjugacy relations for models such as LDA, Gaussian Mixture
Models or Hidden Markov Models, but it suffers from the same
shortcomings as implementations of BUGS when deeper mathematics are
required to discover a conjugacy relation (as would be the case for
instance for a non-linear regression). In the cases where a conjugacy
relation cannot be found, the compiler will (like BUGS) resort to
using Metropolis-Hastings and therefore exploit the inherent
parallelism of these algorithms.

Finally, note that the rewrite rules are applied deterministically and
the process will always terminate and produce the same
result. Overall, the cost of analysis is negligible compared to the
sampling time for large data sets. Although the rewrite system is
simple, it enables us to use a concise symbolic representation for the
model and thereby scale to large networks.


\subsection{Data-parallel Operations on Distributions}

To produce efficient parallel code, the compiler needs to uncover parallelism,
but we also need to rely on a good library of data-parallel operations for
distributions. For instance, in the case of LDA, there are two steps in which
we need to draw from many Dirichlet distributions in parallel. In the first
case, when drawing the topic distributions for the documents, each thread can
draw one of the $\theta_i$ by generating $K$ Gamma variates and normalizing them
\cite{Gamma}. Since the number of documents is usually very large, this
produces enough parallelism to make full use of the GPU's cores.


\begin{algorithm}[tb]
   \caption{Sampling from K Dirichlet Variates}
   \label{Dircomplex}
\begin{algorithmic}
   \STATE {\bfseries Input:} matrix $a$ of size $k$ by $n$
   \FOR{$i=0$ {\bfseries to} $n-1$ in parallel}
   \FOR{$j=0$ to $k-1$}
   \STATE $v[i,j] \sim {\tt Gamma}(a[i,j])$ 
   \ENDFOR
   \STATE $v \; \times \stackrel{\rightarrow}{1}$ 
   \ENDFOR
   \STATE {\bfseries Output:} matrix $v$
\end{algorithmic}
\end{algorithm}

However, this will not produce sufficient parallelism when drawing the
$\phi_k$, because the number of topics is usually small compared to
the number of cores. Consequently, we use a different procedure
(Algorithm \ref{Dircomplex}). To generate $K$ Dirichlet variates over
$V$ categories with concentration parameters $\alpha_{11}, \dots,
\alpha_{KV}$, we first need to generate a matrix $A$ where $A_{ij}
\sim {\tt Gamma}(\alpha_{ij})$ and then normalize each row of this
matrix.  For sampling the $\theta_i$, we were effectively launching a
thread for each row. Now that the number of columns is much larger
than the number of rows, we launch a thread to generate the gamma
variates for each column, and then separately compute a normalizing
constant for each row by multipying the matrix by an all-ones vector
using CUBLAS.  This is an instance where the two-stage compilation
procedure (Section \ref{Design}) is useful, because the compiler is
able to use information about the relative sizes of $K$ and $V$ to
decide that Algorithm \ref{Dircomplex} will be more efficient than the
simple scheme.

This sort of optimization is not unique to the Dirichlet
distribution. For example, when generating a large number of
multinormal variates by applying a linear transformation to a vector
of normal variates, the strategy for extracting parallelism
may change based on the number of variates to generate, the dimension
of the multinormal, and the number of GPU cores.  We found that
to use the GPU effectively we had to develop the language in concert
with the creation of a library of data-parallel operations on
distributions.

\subsection{Parallelism \& Inference Tradeoffs}

It is difficult to give a cost model for Augur programs. Traditional
approaches are not necessarily appropriate for probabilistic programs
because there are tradeoffs between faster sampling times and
convergence which are not easy to characterize. In particular,
different inference methods may affect the amount of parallelism that
can be exploited in a model. For example, in the case of
multivariate regression, we can use the Metropolis-Hastings sampler
presented above, which lets us sample from all the weights in
parallel. However, we may be better off generating a
Metropolis-Within-Gibbs sampler where the weights are sampled one at a
time. This reduces the amount of exploitable parallelism, but it
may converge faster, and there may still be enough
parallelism in each step of Metropolis-Hastings by evaluating the
likelihood in parallel.

In the Hidden-Markov model, once again, one may try to sample the
state of the Markov chain in parallel using a Metropolis-Hastings
sampler just for these variables. If the HMM is small this may be a
good way to make use of a GPU. Of course, for a large HMM it will be
more effective to sample the states of the Markov chain in sequence,
and in this case there is less parallelism to exploit.


In each of these cases, it is not clear which of these alternatives is
better, and different models may perform best with different settings.
Despite these difficulties, because Augur tries to parallelize
operations over arrays, users can maximize the amount of parallelism
in their models by structuring them so that data and parameters are
stored in large, flattened arrays. In addition, as more options and
inference strategies are added to Augur, users will be able to
experiment with the tradeoffs of different inference methods in a way
that would be too time-consuming to do manually.


\section{Experimental Study}
\label{Experiments}

To assess the runtime performance of the inference code generated by
Augur, we present the results of benchmarks for the two examples
presented throughout the paper and for a Gaussian Mixture Model
(GMM). More detailed information on the experiments can be found in
the supplementary material.

For the multivariate regression and GMM, we compare Augur's
performance to those of two popular languages for statistical
modeling, JAGS~\cite{JAGS} and Stan~\cite{Stan}. JAGS is an
implementation of the BUGS language, and performs inference using
Gibbs sampling, adaptative Metropolis-Hastings, and slice
sampling. Stan uses No-U-Turn sampling, a variant of Hamiltonian Monte
Carlo sampling. In the regression experiment, we configured Augur to
use Metropolis-Hastings\footnote{A better way to do inference in the
  case of the regression would have been for Augur to produce a Gibbs
  sampler, but this is not currently implemented. The conjugacy
  relation for the weights is not just an application of conjugacy
  rules \cite{Neal}.  We could add specific rules for linear
  regressions (which is what JAGS does). However, we leave it for
  future work to make the compiler user extensible.}, while for the
GMM experiments Augur generated a Gibbs sampler.

In addition to JAGS and Stan, for the LDA benchmarks we also compare
Augur to a handwritten CUDA implementation of a Gibbs sampler and a
Scala implementation of a collapsed Gibbs sampler~\cite{PNAS} from the
Factorie library~\cite{Factorie}. The former gives us a reference
comparison for what might be possible for a manually optimized GPU
implementation, while the latter gives a baseline for a Scala
implementation that does not use GPUs.

\subsection{Experimental Setup}

For the linear regression experiment, we used data sets from the UCI
regression repository~\cite{UCI}. The Gaussian Mixture Model
experiments used two synthetic data sets, one generated from 3
clusters, the other from 4 clusters.  For the LDA benchmark, we used a
corpus extracted from the simple English variant of Wikipedia, with
standard stopwords removed.  This gives a corpus with 48556 documents,
a vocabulary size of 37276 words, and approximately 3.3 million
tokens.  From that we sampled 1000 documents to use as a test set,
removing words which appear only in the test set.  To evaluate the
model fit we use the log predictive probability measure
\cite{Comparison} on the test set.

All experiments were run on a single workstation with an Intel Core i7
3770 @3.4GHz CPU, 32 GB RAM, a Samsung 840 SSD, and an NVIDIA Geforce
Titan. The Titan runs on the Kepler architecture.  All
probability values are calculated in double precision. The CPU
performance results using Factorie are calculated using a single
thread, as the multi-threaded samplers are neither stable nor
performant in the tested release. The GPU results use all 896
double-precision ALU cores available in the Titan\footnote{The Titan
  has 2688 single-precision ALU cores, but single precision resulted
  in poor quality inference results, though the speed was greatly
  improved.}.

\subsection{Results}

In general, our results show that once the problem is large enough we
can amortize Augur's startup cost of model compilation to CUDA, {\tt
  nvcc} compilation to a GPU binary, and copying the data to and from
the GPU.  This cost is approximately 10 seconds on average across all
our experiments. After this point Augur scales to larger numbers of
samples in shorter runtimes than comparable systems. More experimental
detail is in the supplementary material.

Our experiments with regression show that Augur's inference is similar
to JAGS in runtime and performance, and better than Stan. This is not
a surprise because the regressions have few random variables and the
data sets are relatively small, and so making use of the GPU is not
justified (except maybe for much larger data sets). However, the
results are very different for models with latent variables where the
number of variables grows with the data set.

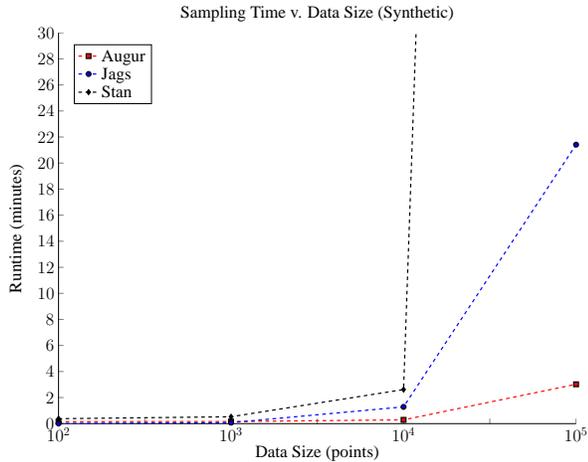
\begin{figure}[t]
\vskip 0.2in
\begin{center}

%
%
\begin{tikzpicture}[scale=0.45,font=\Large]

\begin{axis}[%
width=6.01947782842934in,
height=4.54928278277609in,
scale only axis,
xmin=2,
xmax=5,
xtick={2,2.5,3,3.5,4,4.5,5},
xticklabels={$10^2$,,$10^3$,,$10^4$,,$10^5$},
xlabel={Data Size (points)},
ymin=0,
ymax=30,
ylabel={Runtime (minutes)},
title={Sampling Time v. Data Size (Synthetic)},
axis x line*=bottom,
axis y line*=left,
legend style={at={(0.03,0.97)},anchor=north west,draw=black,fill=white,legend cell align=left}
]
\addplot [
color=red,
dashed,
line width=1.0pt,
mark=square*,
mark options={solid,fill=red,draw=black}
]
table[row sep=crcr]{
2 0.1556\\
3 0.16055\\
4 0.28955\\
5 3.01873333333333\\
};
\addlegendentry{Augur};

\addplot [
color=blue,
dashed,
line width=1.0pt,
mark=*,
mark options={solid,fill=blue,draw=black}
]
table[row sep=crcr]{
2 0.0065\\
3 0.07615\\
4 1.28528333333333\\
5 21.41305\\
};
\addlegendentry{Jags};

\addplot [
color=black,
dashed,
line width=1.0pt,
mark=diamond*,
mark options={solid,fill=black,draw=black}
]
table[row sep=crcr]{
2 0.377\\
3 0.528016666666667\\
4 2.61363333333333\\
5 380.539016666667\\
};
\addlegendentry{Stan};

\end{axis}
\end{tikzpicture}%

\caption{Evolution of runtime to draw a thousand samples for varying
  data set sizes for a Gaussian Mixture Model. Stan's last data point is
  cropped, it took 380 minutes.}
\label{GMM}
\end{center}
\vskip -0.2in 
\end{figure} 

For instance, using the GMM example, we show (Figure \ref{GMM}) that
Augur scales better than JAGS and Stan. For a hundred thousand data
points, Augur draws a thousand samples in about 3 minutes whereas JAGS
needs more than 21 minutes and Stan requires more than 6 hours. Each
system found the correct means and variances for the clusters, the aim
here is to measure the scaling of runtime with problem size.

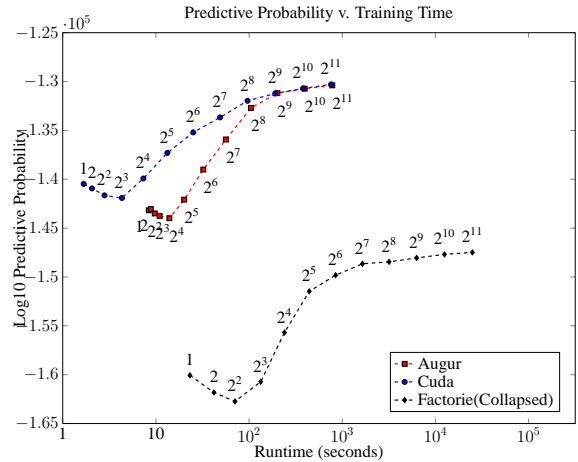
\begin{figure}[t]
\vskip 0.2in
\begin{center}

%
%
\begin{tikzpicture}[scale=0.45,font=\Large]

\begin{axis}[%
width=5.96122481718648in,
height=4.54928278277609in,
scale only axis,
xmin=0,
xmax=5.5,
xtick={0,1,2,3,4,5},
xticklabels={1,10,$10^2$,$10^3$,$10^4$,$10^5$},
xlabel={Runtime (seconds)},
ymin=-165000,
ymax=-125000,
ylabel={Log10 Predictive Probability},
title={Predictive Probability v. Training Time},
legend style={at={(0.97,0.03)},anchor=south east,draw=black,fill=white,legend cell align=left}
]
\addplot [
color=red,
dashed,
line width=1.0pt,
mark=square*,
mark options={solid,fill=red,draw=black}
]
table[row sep=crcr]{
0.9237619608287 -143177.770499382\\
0.94546858513182 -143045.310682402\\
0.988112840268352 -143497.411238063\\
1.04099769242349 -143765.805227879\\
1.14457420760962 -143983.821057795\\
1.30319605742049 -142086.388466359\\
1.50839503313305 -139008.109178629\\
1.75120209458835 -135931.132774344\\
2.02031984035731 -132695.638884165\\
2.30377904659553 -131184.294087142\\
2.59632213868033 -130729.587764589\\
2.892912310462 -130385.19224044\\
};
\addlegendentry{Augur};

\node[above, right, inner sep=0mm, text=black, yshift=4mm]
at (axis cs:0.7737619608287,-146177.770499382,0) {1};
\node[above, right, inner sep=0mm, text=black, yshift=4mm]
at (axis cs:0.82546858513182,-146045.310682402,0) {2};
\node[above, right, inner sep=0mm, text=black, yshift=4mm]
at (axis cs:0.908112840268352,-146497.411238063,0) {$\text{2}^\text{2}$};
\node[above, right, inner sep=0mm, text=black, yshift=4mm]
at (axis cs:0.99099769242349,-146765.805227879,0) {$\text{2}^\text{3}$};
\node[above, right, inner sep=0mm, text=black, yshift=4mm]
at (axis cs:1.14457420760962,-146983.821057795,0) {$\text{2}^\text{4}$};
\node[above, right, inner sep=0mm, text=black, yshift=4mm]
at (axis cs:1.30319605742049,-145086.388466359,0) {$\text{2}^\text{5}$};
\node[above, right, inner sep=0mm, text=black, yshift=4mm]
at (axis cs:1.50839503313305,-142008.109178629,0) {$\text{2}^\text{6}$};
\node[above, right, inner sep=0mm, text=black, yshift=4mm]
at (axis cs:1.75120209458835,-138931.132774344,0) {$\text{2}^\text{7}$};
\node[above, right, inner sep=0mm, text=black, yshift=4mm]
at (axis cs:2.02031984035731,-135695.638884165,0) {$\text{2}^\text{8}$};
\node[above, right, inner sep=0mm, text=black, yshift=4mm]
at (axis cs:2.30377904659553,-134184.294087142,0) {$\text{2}^\text{9}$};
\node[above, right, inner sep=0mm, text=black, yshift=4mm]
at (axis cs:2.59632213868033,-133729.587764589,0) {$\text{2}^{\text{10}}$};
\node[above, right, inner sep=0mm, text=black, yshift=4mm]
at (axis cs:2.892912310462,-133385.19224044,0) {$\text{2}^{\text{11}}$};
\addplot [
color=blue,
dashed,
line width=1.0pt,
mark=*,
mark options={solid,fill=blue,draw=black}
]
table[row sep=crcr]{
0.225309281725863 -140478.630294353\\
0.313867220369153 -140937.245267243\\
0.44870631990508 -141648.619628601\\
0.635483746814912 -141911.367790152\\
0.865696059916071 -139919.259001662\\
1.123198075032 -137303.069042677\\
1.40088321554836 -135200.649455783\\
1.68922003726384 -133667.155640183\\
1.98376156028616 -131966.892743532\\
2.28062389151291 -131219.906234658\\
2.57977216766427 -130700.924328784\\
2.88050490494437 -130285.304509602\\
};
\addlegendentry{Cuda};

\node[above, inner sep=0mm, text=black]
at (axis cs:0.225309281725863,-139478.630294353,0) {1};
\node[above, inner sep=0mm, text=black]
at (axis cs:0.313867220369153,-139937.245267243,0) {2};
\node[above, inner sep=0mm, text=black]
at (axis cs:0.44870631990508,-140648.619628601,0) {$\text{2}^\text{2}$};
\node[above, inner sep=0mm, text=black]
at (axis cs:0.635483746814912,-140911.367790152,0) {$\text{2}^\text{3}$};
\node[above, inner sep=0mm, text=black]
at (axis cs:0.865696059916071,-138919.259001662,0) {$\text{2}^\text{4}$};
\node[above, inner sep=0mm, text=black]
at (axis cs:1.123198075032,-136303.069042677,0) {$\text{2}^\text{5}$};
\node[above, inner sep=0mm, text=black]
at (axis cs:1.40088321554836,-134200.649455783,0) {$\text{2}^\text{6}$};
\node[above, inner sep=0mm, text=black]
at (axis cs:1.68922003726384,-132667.155640183,0) {$\text{2}^\text{7}$};
\node[above, inner sep=0mm, text=black]
at (axis cs:1.98376156028616,-130966.892743532,0) {$\text{2}^\text{8}$};
\node[above, inner sep=0mm, text=black]
at (axis cs:2.28062389151291,-130219.906234658,0) {$\text{2}^\text{9}$};
\node[above, inner sep=0mm, text=black]
at (axis cs:2.57977216766427,-129700.924328784,0) {$\text{2}^{\text{10}}$};
\node[above, inner sep=0mm, text=black]
at (axis cs:2.88050490494437,-129285.304509602,0) {$\text{2}^{\text{11}}$};
\addplot [
color=black,
dashed,
line width=1.0pt,
mark=diamond*,
mark options={solid,fill=black,draw=black}
]
table[row sep=crcr]{
1.3647385550554 -160067.4829006\\
1.62304243424638 -161818.558251634\\
1.84874318189568 -162737.091080859\\
2.12505815117207 -160723.267568274\\
2.37945056057798 -155691.097406461\\
2.64617818746451 -151472.360809252\\
2.92862113529216 -149815.527360791\\
3.2173207241047 -148649.012382399\\
3.50284482505623 -148452.711276579\\
3.79836539499743 -148049.251702891\\
4.09667960210448 -147686.18151602\\
4.39646176305844 -147485.537465381\\
};
\addlegendentry{Factorie(Collapsed)};

\node[above, inner sep=0mm, text=black]
at (axis cs:1.3647385550554,-159067.4829006,0) {1};
\node[above, inner sep=0mm, text=black]
at (axis cs:1.62304243424638,-160818.558251634,0) {2};
\node[above, inner sep=0mm, text=black]
at (axis cs:1.84874318189568,-161737.091080859,0) {$\text{2}^\text{2}$};
\node[above, inner sep=0mm, text=black]
at (axis cs:2.12505815117207,-159723.267568274,0) {$\text{2}^\text{3}$};
\node[above, inner sep=0mm, text=black]
at (axis cs:2.37945056057798,-154691.097406461,0) {$\text{2}^\text{4}$};
\node[above, inner sep=0mm, text=black]
at (axis cs:2.64617818746451,-150472.360809252,0) {$\text{2}^\text{5}$};
\node[above, inner sep=0mm, text=black]
at (axis cs:2.92862113529216,-148815.527360791,0) {$\text{2}^\text{6}$};
\node[above, inner sep=0mm, text=black]
at (axis cs:3.2173207241047,-147649.012382399,0) {$\text{2}^\text{7}$};
\node[above, inner sep=0mm, text=black]
at (axis cs:3.50284482505623,-147452.711276579,0) {$\text{2}^\text{8}$};
\node[above, inner sep=0mm, text=black]
at (axis cs:3.79836539499743,-147049.251702891,0) {$\text{2}^\text{9}$};
\node[above, inner sep=0mm, text=black]
at (axis cs:4.09667960210448,-146686.18151602,0) {$\text{2}^{\text{10}}$};
\node[above, inner sep=0mm, text=black]
at (axis cs:4.39646176305844,-146485.537465381,0) {$\text{2}^{\text{11}}$};
\end{axis}
\end{tikzpicture}%

\caption{Evolution of the predictive probability over time for up to
  2048 samples and for three implementations of LDA inference: Augur,
  hand-written CUDA, Factorie's Collapsed Gibbs.}
\label{Long}
\end{center}
\vskip -0.2in 
\end{figure} 

Results from the LDA experiment are presented in Figure \ref{Long} and
use predictive probability to compare convergence over time. We
compute the predictive probability and record the time after drawing
$2^i$ samples, for $i$ ranging from 0 to 11 inclusive. The time is
reported in seconds. It takes Augur 8.1 seconds to draw its first
sample for LDA. The inference for Augur is very close to that of the
hand-written CUDA implementation, and much faster than the Factorie
collapsed Gibbs sampler. Indeed, it takes 6.7 more hours for the
collapsed LDA implementation to draw 2048 samples than it does for
Augur.


We also implemented LDA in the JAGS and Stan systems but they run into
scalability issues. The Stan version of LDA uses 55 gigabytes of RAM
but failed to draw a second sample given a week of computation
time. Unfortunately, we could not apply JAGS because it requires more
than 128 gigabytes of memory. In comparison, Augur uses less than a
gigabyte of memory for this experiment.

\section{Related Work}
\label{Related}

To our knowledge, BUGS \cite{BUGS} was the first probabilistic
programming language. Interestingly, most of the key concepts of
probabilistic programming already appeared in the first paper to
introduce BUGS \cite{BUGSOriginal}. Since then, research in
probabilistic programming languages has been focused in two
directions: improving performance and scalability through better
inference generation; and, increasing expressiveness and building the
foundations of a universal probabilistic programming language. These
two directions are useful criteria to compare probabilistic
programming languages.

In terms of language expressiveness, Augur is currently limited to the
specification of Bayesian networks. It is possible to extend this
language (e.g., non-parametric models) or to add new modeling
languages (e.g., Markov random field), but our current focus
is on improving the inference generation. That is in contrast with
languages like Hansei \cite{Hansei}, Odds \cite{Odds}, Stochastic Lisp
\cite{Slisp} and Ibal \cite{Ibal} which focus on increasing
expressiveness, at the expense of performance.  However, as Augur is
embedded in the Scala programming language, we have access to the wide
variety of libraries on the JVM platform and benefit from Scala
tools. Augur, like Stan \cite{Stan} and BUGS \cite{BUGSOriginal,BUGS}
is a domain specific probabilistic language for Bayesian networks, but
it is embedded in such a way that it has a very good integration with
the rest of Scala, which is crucial to software projects where data
analysis is only one component of a larger artifact.


Augur is not the only system designed for scalability and
performance. It is also the case of Dimple \cite{Dimple}, Factorie
\cite{Factorie}, Infer.net \cite{Infer} and Figaro \cite{Figaro1,
  Figaro2}, and the latest versions of Church \cite{Church}.  Dimple
focuses on performance using specialized inference hardware, though it
does provide an interface for CPU code. Factorie mainly focuses on
undirected networks, and is a Scala library rather than a DSL (unlike
all the other systems mentioned). It has multiple inference backends,
and aims to be a general purpose machine learning package.  Infer.net
is the system most similar to Augur, in that it has a two phase
compilation approach, though it is based around variational methods. A
block Gibbs sampler exists but is only functional on a subset of the
models.  Figaro focuses on a different set of inference techniques,
including techniques which use exact inference in discrete spaces
(they also have Metropolis-Hastings inference). Church provides the
ability to mix different inference algorithms and has some parallel
capability, but it is focused on task-parallelism for multicores
rather than on data-parallelism for parallel architectures. GraphLab
\cite{GRAPHLAB} is another framework for parallel machine learning
which is more focused on multiprocessor and distributed computing than
on the kind of data-parallelism available on GPUs.
The key difference between Augur and these other languages is the
systematic generation of data-parallel algorithms for large numbers of
cores (i.e., thousands) on generally available GPU hardware, and the
use of a symbolic representation of the model in the compiler.







\section{Conclusion}
\label{Conclusion}

We find that it is possible to \emph{automatically} generate parallel
MCMC-based inference algorithms, and it is also possible to extract
sufficient parallelism to saturate a modern GPU with thousands of
cores.  Our compiler achieves this with no extra information beyond
that which is normally encoded in a graphical model description and
uses a symbolic representation that allows scaling to large models
(particularly for latent variable models such as LDA). It also makes
it easy to run different inference algorithms and evaluate the
tradeoffs between convergence and sampling time. The generated
inference code is competitive in terms of sample quality with other
probabilistic programming systems, and for large problems generates
samples much more quickly.

\bibliography{ref}
\bibliographystyle{icml2014}

\appendix

\section{Examples of Model Specification}

We present a few examples of model specifications in Augur, covering
three important topics in machine learning: regression
(\ref{Polynomial}), clustering (\ref{Categorical},
\ref{Gaussian},\ref{secHMM}), and classification (\ref{Naive}). Our goal
is to show how a few popular models can be programmed in Augur. For
each of these examples, we first describe the support of the model,
and then sketch the generative process, relating the most complex
parts of the program to their usual mathematical notation.

\subsection{Univariate polynomial regression}
\label{Polynomial}

Our first example model is for univariate polynomial regression
(Figure \ref{PR}). The model's support is composed of the array {\tt w}
for the weights of each mononomial, {\tt x} for the domain data points
and {\tt y} for their image. The parameters of the model are: {\tt N},
the dataset size and {\tt M}, the order of the polynomial. For simplicity, this
example assumes that the domain of {\tt x} ranges from 0 to 2.

The generative process is: We first independently draw each of the
{\tt M} weights, $w_i \sim N(0,1)$, then draw $(x,y)$ as follows:
\begin{align}
    x_j &\sim \mbox{Uniform}(0,2) \\
    y_j &\sim N(\sum\limits_i^M w_ix_j^i,1).
\end{align}
For simplicity, the model is presented with many ``hardwired''
parameters, but it is possible to parameterize the model to control the
noise level, or the domain of {\tt x}.

\begin{figure*}
\vskip 0.2in
\begin{center}
\begin{minipage}{0.9\textwidth}
\begin{lstlisting}
object UnivariatePolynomialRegression {

  import scala.math._

  class sig(var w: Array[Double], var x: Array[Double], var y: Array[Double])       
  
  val model = bayes {
    (N: Int, M: Int) => {        
        
    val w = Gaussian(0,1).sample(M)
    val x = Uniform(0,2).sample(N)
    val bias = Gaussian(0,1).sample
    val y = for(i <- 1 to N) {
               val monomials = for (j <- 1 to M) yield { w(j) * pow(x(i),j) }
               Gaussian((monomials.sum) + bias, 1).sample()       
            }

    observe(x, y)
    }
  }
}
\end{lstlisting}
\end{minipage}
\caption{Specification of a univariate polynomial regression}
\label{PR}
\end{center}
\vskip -0.2in
\end{figure*}

\subsection{Categorical mixture}
\label{Categorical}

The third example is a categorical mixture model (Figure
\ref{categorical}). The model's support is composed of an array {\tt
  z} for the cluster selection, {\tt x} for the data points that we
draw, {\tt theta} for the priors of the categorical that represents
the data, and {\tt phi} for the prior of the indicator variable. The
parameters of the model are: {\tt N} data size, {\tt K} number of
clusters, and {\tt V} for the vocabulary size.

The generative process is: For each of the {\tt N} data
points we want to draw, we select a cluster {\tt z} according to their
distribution {\tt phi} and then draw from the categorical with
distribution given by {\tt theta(z)}.

\begin{figure*}
\vskip 0.2in
\begin{center}
\begin{minipage}{0.9\textwidth}
\begin{lstlisting}
object CategoricalMixture {
  class sig(var z: Array[Int], var x: Array[Int], var theta: Array[Double], var phi: Array[Double])
  val model = bayes {
    (N: Int, K: Int, V: Int) => {
        
      val alpha = vector(V,0.5)
      val beta = vector(K,0.5)
        
      val theta = Dirichlet(V,alpha).sample(K)
      val phi = Dirichlet(K,beta).sample()
        
      val x = for(i <- 1 to N) {
      		  val z = Categorical(K, phi).sample()
      		  Categorical(N,theta(z)).sample()
      		}
      observe(x)
    }
  }
}
\end{lstlisting}
\end{minipage}
\caption{Specification of a categorical mixture model}
\label{categorical}
\end{center}
\vskip -0.2in
\end{figure*}

\subsection{Gaussian Mixture Model}
\label{Gaussian}

The fourth example is a univariate Gaussian mixture model (Figure
\ref{GMM}). The model's support is composed of an array {\tt z} for
the cluster selection, {\tt x} for the data points that we draw, {\tt
  mu} for the priors over the cluster means, {\tt sigma} for the
priors of the cluster variances, and {\tt phi} for the prior of
the indicator variable. The parameters of the model are: {\tt N} data
size, {\tt K} number of clusters.

The generative process is: For each of the {\tt N} data
points we want to draw, we select a cluster {\tt z} according to their
distribution {\tt pi} and then draw from the Gaussian centered at {\tt
  mu(z)} and of standard deviation {\tt sigma(z)}.

\begin{figure*}
\vskip 0.2in
\begin{center}
\begin{minipage}{0.9\textwidth}
\begin{lstlisting}
object GaussianMixture {

  class sig(var z: Array[Int], var x: Array[Double], var mu: Array[Double], var sigma: Array[Double], var phi: Array[Double])

  val model = bayes {
    (N: Int, K: Int, V: Int) => {
      
    val alpha = vector(V,0.1)  
      
    val phi = Dirichlet(V,alpha).sample()  
    val mu = Gaussian(0,10).sample(K)
    val sigma = InverseGamma(1,1).sample(K)   
    
    val x = for(i <- 1 to N) {
               val z = Categorical(K, phi).sample()
               Gaussian(mu(z), sigma(z)).sample()
            }

      observe(x)
    }
  }
}
\end{lstlisting}
\end{minipage}
\caption{Specification of a Gaussian mixture model}
\label{GMM}
\end{center}
\vskip -0.2in
\end{figure*}
\subsection{Naive bayes classifier}
\label{Naive}

The fifth example is a binary naive Bayes classifier (Figure
\ref{NBC}). The support is composed of an array {\tt c} for the class
and an array {\tt f} for the features, {\tt pC} the prior on the
positive class, and {\tt pFgivenC} an array for the probability of
each binary feature given the class.  The hyperparameters of the model
are: {\tt N} the number of data points, {\tt K} the number of features
and.  The features form a 2-dimensional matrix but again the user has
to ``flatten'' the matrix into an array.

The generative process is: First we draw the probability
of an event being in one class or the other as {\tt pC}. We use {\tt
  pC} has the parameter to decide for each event in which class it
falls ({\tt c}). Then, for each feature, we draw the probability of
the feature occurring, {\tt pFgivenC}, depending on whether the event
is in the class or not. Finally, we draw the features {\tt f} for each
event.

\begin{figure*}
\vskip 0.2in
\begin{center}
\begin{minipage}{0.9\textwidth}
\begin{lstlisting}
object NaiveBayesClassifier {

  class sig(var c: Array[Int], var f: Array[Int], var pC: Double, var pFgivenC: Array[Double])

  val model = bayes {
    (N: Int, K: Int) => {
        
    val pC = Beta(0.5,0.5).sample()  
    val c = Bernoulli(pC).sample(N)

    val pFgivenC = Beta(0.5,0.5).sample(K*2)
        
    val f = for(i <- 1 to N) {
               for(j <- 1 to K) {
                  Bernoulli(pFgivenC(j * 2 + c(i))).sample()
               }
            }
        
    observe(f, c)
    }
  }
}
\end{lstlisting}
\end{minipage}
\caption{Specification of a naive Bayes classifier}
\label{NBC}
\end{center}
\vskip -0.2in
\end{figure*}

\subsection{Hidden Markov Model}
\label{secHMM}

\begin{figure*}
\vskip 0.2in
\begin{center}
\begin{minipage}{0.9\textwidth}
\begin{lstlisting}
object HiddenMarkovModel {
  class sig(var flips: Array[Int], var bias: Array[Double], var transition_matrix: Array[Double], var MC_states: Array[Int])
    
  val model = bayes {
    (N: Int, number_states: Int) => {
      val v = vector(number_states,0.1)
      val transition_matrix = Dirichlet(number_states,v).sample(number_states)          
      val bias = Beta(1.0,1.0).sample(number_states)
        
      val MC_states: IndexedSeq[Int] = for (i <- 1 to N) yield Categorical(number_states,transition_matrix(MC_states(max(0, i-1)))).sample()  
        
      val flips = for (i <- 1 to N) Bernoulli(bias(MC_states(i))).sample()
      
      observe(flips)
      }    
    }
  }
\end{lstlisting}
\end{minipage}
\caption{Specification of a Hidden Markov Model}
\label{HMM}
\end{center}
\vskip -0.2in
\end{figure*}

The sixth example is a hidden Markov model (Figure \ref{HMM}) where
the observation are the result of coin flips. The support is composed
of the result of the coin flips {\tt flips}, the priors for each of
the coins {\tt bias}, the transition matrix to decide how to change
coin {\tt transition\_matrix}, and the states of the Markov chain that
indicates which coin is being used for the flip {\tt MC\_states}. The
two parameters of the model are the size of the data {\tt N}, and the
number of coins being used {\tt number\_states}.

The generative process is: draw a transition matrix for the Markov
chain, a bias for each of the coins, decide what coin is to be used in
each state, using the transition matrix, and based on this, flip the
coin that should be used for each state.

\section{Simple Data-Parallel Sampling from M Dirichlet Distributions}

The algorithm in \ref{simpledir} presents a simple way to draw from a
number of Dirichlet distributions in parallel on a GPU. It works well
if the number $M$ is very large. On the contrary, it is a bottleneck
if $M$ is small or much lesser than the dimension of the Dirichlet
distributions.

\begin{algorithm}[tb]
   \caption{Sampling from Dirichlet($\alpha$) M times}
   \label{simpledir}
\begin{algorithmic}
   \STATE {\bfseries Input:} array $\alpha$ of size $n$
   \FOR{M documents in parallel}
   \FOR{$i=0$ {\bfseries to} $n-1$} 
   \STATE $v[i] \sim {\tt Gamma}(a[i])$
   \ENDFOR
   \STATE $s = \sum\limits_0^{n-1} a[i]$ in parallel
   \FOR{$i=0$ {\bfseries to} $n-1$ in parallel} 
   \STATE $v[i] = \frac{v[i]}{s}$
   \ENDFOR    
   \ENDFOR
   \STATE {\bfseries Output:} array $v$
\end{algorithmic}
\end{algorithm}

\section{Experimental study}

This section contains additional data from the benchmarks.

\subsection{Multivariate Regression}

In our regression experiment, we compare Augur against two other
models, one implemented in Jags (\ref{JAGS}) and one in Stan
(\ref{STAN}). These models are both based upon the BMLR code developed
by Kruschke \cite{kruschke2010doing}. Each system uses the same priors
and hyperparameters.

\begin{figure*}
\vskip 0.2in
\begin{center}
\begin{minipage}{0.9\textwidth}
\begin{lstlisting}
model {
    for( i in 1:N ) {
      y[i] ~ dnorm( y.hat[i] , 1/tau )
      y.hat[i] <- b0 +  inprod( b[1:nPred] , x[i,1:nPred] )
    }
    tau ~ dgamma( 1 , 1 )
    b0 ~ dnorm( 0 , 0.01 )
    for ( j in 1:nPred ) {
      b[j] ~ dnorm( 0 , 0.01 )
    }
  }
\end{lstlisting}
\end{minipage}
\caption{Multivariate Regression in Jags}
\label{JAGS}
\end{center}
\vskip -0.2in
\end{figure*}

\begin{figure*}
\vskip 0.2in
\begin{center}
\begin{minipage}{0.9\textwidth}
\begin{lstlisting}
data {
  int<lower=0> nPred;
  int<lower=0> nData;
  real y[nData];
  matrix[nData,nPred] x;
  vector[nData] b0vec;
}
parameters {
  real b0;
  vector[nPred] b;
  real<lower=0> tau;
}
transformed parameters {
  vector[nData] mu;
  vector[nData] offset;
  offset <- b0vec * b0;
  mu <- x * b + offset;
}
model {
  b0 ~ normal(0,10);
  tau ~ gamma(1,1);
  for (d in 1:nPred)
    b[d] ~ normal(0,10);
  y ~ normal(mu, 1/sqrt(tau));
}
\end{lstlisting}
\end{minipage}
\caption{Multivariate Regression in Stan}
\label{STAN}
\end{center}
\vskip -0.2in
\end{figure*}

The regression experimental protocol was as follows: each dataset had 10 90\%/10\% train/test
splits generated, and each dataset was tested using 10 different random initialisations across
each of the train/test splits. Then the number of samples was varied between 100, 200, 500, 
1000, 2000, 5000\footnote{Stan and JAGS had a burn in of an additional 50\% samples to allow for the 
adaptive tuning of the samplers, without these extra samples for adaptation the performance of 
both of them was poor. Augur's Metropolis-Hastings algorithm does not use such tuning.}. 
This gives a total of 600 runs of each system on each dataset.
The presented figures average across both the random seeds and the train/test splits
to produce one point per number of samples. We then plot average RMSE on the test sets 
against average runtime. 

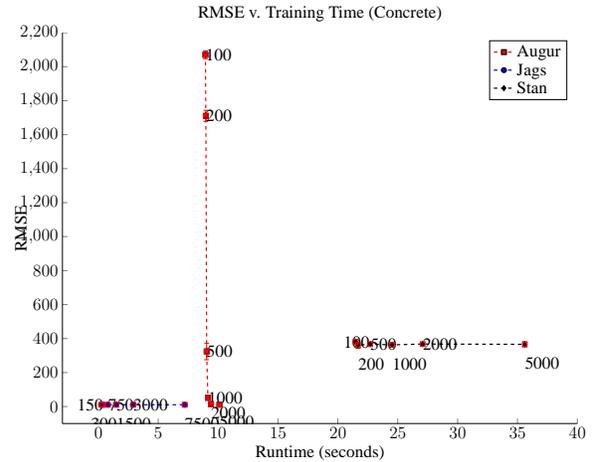
\begin{figure}[t]
\vskip 0.2in
\begin{center}

%
%
\begin{tikzpicture}[scale=0.45,font=\Large]

\begin{axis}[%
width=5.98896434634975in,
height=4.54928278277609in,
scale only axis,
xmin=-3,
xmax=40,
xlabel={Runtime (seconds)},
ymin=-100,
ymax=2200,
ylabel={RMSE},
title={RMSE v. Training Time (Concrete)},
axis x line*=bottom,
axis y line*=left,
legend style={draw=black,fill=white,legend cell align=left}
]
\addplot [
color=red,
dashed,
line width=1.0pt,
mark=square*,
mark options={solid,fill=red,draw=black}
]
table[row sep=crcr]{
8.94058 2070.0877711248\\
8.97471 1710.7017260628\\
9.04423 324.192363760309\\
9.14719 51.086415129755\\
9.40565 14.280925839489\\
10.11101 10.9607185663299\\
};
\addlegendentry{Augur};

\addplot [
color=red,
only marks,
mark=square,
mark options={solid},
forget plot
]
plot [error bars/.cd, y dir = both, y explicit]
coordinates{
(8.94058,2070.0877711248) +- (0.0,23.0553972264806)(8.97471,1710.7017260628) +- (0.0,32.4976652235327)(9.04423,324.192363760309) +- (0.0,48.5244286397038)(9.14719,51.086415129755) +- (0.0,3.03983825676528)(9.40565,14.280925839489) +- (0.0,0.583018250454689)(10.11101,10.9607185663299) +- (0.0,0.0918402036699657)};
\node[above, right, inner sep=0mm, text=black]
at (axis cs:8.94058,2070.0877711248,0) {100};
\node[above, right, inner sep=0mm, text=black]
at (axis cs:8.97471,1710.7017260628,0) {200};
\node[above, right, inner sep=0mm, text=black]
at (axis cs:9.04423,324.192363760309,0) {500};
\node[above, right, inner sep=0mm, text=black]
at (axis cs:9.14719,51.086415129755,0) {1000};
\node[above, right, inner sep=0mm, text=black]
at (axis cs:9.40565,-35.719074160511,0) {2000};
\node[above, right, inner sep=0mm, text=black]
at (axis cs:10.11101,-89.0392814336701,0) {5000};
\addplot [
color=blue,
dashed,
line width=1.0pt,
mark=*,
mark options={solid,fill=blue,draw=black}
]
table[row sep=crcr]{
0.23439 10.4653599501266\\
0.36932 10.4696707568023\\
0.79404 10.4470027517381\\
1.50387 10.4790923788256\\
2.92205 10.45358370255\\
7.21352 10.4630097043902\\
};
\addlegendentry{Jags};

\addplot [
color=red,
only marks,
mark=square,
mark options={solid},
forget plot
]
plot [error bars/.cd, y dir = both, y explicit]
coordinates{
(0.23439,10.4653599501266) +- (0.0,0.0263796832596963)(0.36932,10.4696707568023) +- (0.0,0.0425428672748236)(0.79404,10.4470027517381) +- (0.0,0.0274979733698597)(1.50387,10.4790923788256) +- (0.0,0.0342936119593737)(2.92205,10.45358370255) +- (0.0,0.0188391244444954)(7.21352,10.4630097043902) +- (0.0,0.0348319755709134)};
\node[above, right, inner sep=0mm, text=black]
at (axis cs:-1.76561,10.4653599501266,0) {150};
\node[above, right, inner sep=0mm, text=black]
at (axis cs:-0.63068,-99.5303292431977,0) {300};
\node[above, right, inner sep=0mm, text=black]
at (axis cs:0.79404,10.4470027517381,0) {750};
\node[above, right, inner sep=0mm, text=black]
at (axis cs:1.50387,-99.5209076211744,0) {1500};
\node[above, right, inner sep=0mm, text=black]
at (axis cs:2.92205,10.45358370255,0) {3000};
\node[above, right, inner sep=0mm, text=black]
at (axis cs:7.21352,-99.5369902956098,0) {7500};
\addplot [
color=black,
dashed,
line width=1.0pt,
mark=diamond*,
mark options={solid,fill=black,draw=black}
]
table[row sep=crcr]{
21.48075 380.9324855113\\
21.68493 360.8740294665\\
22.70501 367.16442607909\\
24.53191 362.99435377175\\
27.07762 367.8326180035\\
35.60986 366.15280036216\\
};
\addlegendentry{Stan};

\addplot [
color=red,
only marks,
mark=square,
mark options={solid},
forget plot
]
plot [error bars/.cd, y dir = both, y explicit]
coordinates{
(21.48075,380.9324855113) +- (0.0,13.1496517386365)(21.68493,360.8740294665) +- (0.0,18.1511176733858)(22.70501,367.16442607909) +- (0.0,8.92815514815925)(24.53191,362.99435377175) +- (0.0,11.8339705001759)(27.07762,367.8326180035) +- (0.0,7.43204855111952)(35.60986,366.15280036216) +- (0.0,15.037195350234)};
\node[above, right, inner sep=0mm, text=black]
at (axis cs:20.48075,380.9324855113,0) {100};
\node[above, right, inner sep=0mm, text=black]
at (axis cs:21.68493,250.8740294665,0) {200};
\node[above, right, inner sep=0mm, text=black]
at (axis cs:22.70501,367.16442607909,0) {500};
\node[above, right, inner sep=0mm, text=black]
at (axis cs:24.53191,252.99435377175,0) {1000};
\node[above, right, inner sep=0mm, text=black]
at (axis cs:27.07762,367.8326180035,0) {2000};
\node[above, right, inner sep=0mm, text=black]
at (axis cs:35.60986,256.15280036216,0) {5000};
\end{axis}
\end{tikzpicture}%

\caption{Result of Multivariate Regression on the Concrete Compressive
Strength data set.}
\label{figConcrete}
\end{center}
\vskip -0.2in
\end{figure}

In figures \ref{figConcrete}, \ref{figRed}, \ref{figWhite} and \ref{figYacht}
we present results on the Concrete compressive, winequality-red, winequality-white and Yacht Hydrodynamics
datasets from the UCI repository \cite{UCI}. In Concrete and Yacht we present results from JAGS,
Stan and Augur. On the winequality experiments we only present results from JAGS and Augur due to machine
time constraints (Stan's runtime was too high to perform sufficiently many experiments on larger datasets). 
JAGS is using a Gibbs sampler for the weights and the bias, and uses a slice sampler
for the variance of the noise. Augur uses random walk Metropolis-Hastings, and Stan is using the No-U-Turn
variant of Hamiltonian Monte Carlo. We can see that Augur has a startup cost of about 10 seconds, and Stan has a 
startup cost of about 20 seconds. After that point Augur can draw samples more quickly than both Stan and
JAGS, though due to JAGS's low startup time (<<1 second) it is only on large datasets with many samples that Augur provides
a speedup.

\begin{figure}[t]
\vskip 0.2in
\begin{center}

%
%
\begin{tikzpicture}[scale=0.45,font=\Large]

\begin{axis}[%
width=6.01947782842934in,
height=4.54928278277609in,
clip=false,
scale only axis,
xmin=0,
xmax=15,
xlabel={Runtime (seconds)},
ymin=-3,
ymax=85,
ylabel={RMSE},
title={RMSE v. Training Time (winequality-red)},
axis x line*=bottom,
axis y line*=left,
legend style={draw=black,fill=white,legend cell align=left}
]
\addplot [
color=red,
dashed,
line width=1.0pt,
mark=square*,
mark options={solid,fill=red,draw=black}
]
table[row sep=crcr]{
8.98053 79.192214370473\\
9.00214 59.695728316981\\
9.12684 11.5103793672049\\
9.29728 1.33082306305847\\
9.63395 0.89270796136464\\
10.65902 0.76436486426649\\
};
\addlegendentry{Augur};

\addplot [
color=red,
only marks,
mark=square,
mark options={solid},
forget plot
]
plot [error bars/.cd, y dir = both, y explicit]
coordinates{
(8.98053,79.192214370473) +- (0.0,1.10580650680981)(9.00214,59.695728316981) +- (0.0,1.98106722738443)(9.12684,11.5103793672049) +- (0.0,0.718400185075074)(9.29728,1.33082306305847) +- (0.0,0.0832877642677707)(9.63395,0.89270796136464) +- (0.0,0.0221679564468137)(10.65902,0.76436486426649) +- (0.0,0.00740635351265218)};
\node[below, right, inner sep=0mm, text=black]
at (axis cs:8.98053,79.192214370473,0) {100};
\node[below, right, inner sep=0mm, text=black]
at (axis cs:9.00214,59.695728316981,0) {200};
\node[below, right, inner sep=0mm, text=black]
at (axis cs:9.12684,11.5103793672049,0) {500};
\node[below, right, inner sep=0mm, text=black]
at (axis cs:9.29728,6.33082306305847,0) {1000};
\node[below, right, inner sep=0mm, text=black]
at (axis cs:9.63395,0.89270796136464,0) {2000};
\node[below, right, inner sep=0mm, text=black]
at (axis cs:10.65902,0.76436486426649,0) {5000};
\addplot [
color=blue,
dashed,
line width=1.0pt,
mark=*,
mark options={solid,fill=blue,draw=black}
]
table[row sep=crcr]{
0.45066 0.65218035294639\\
0.71125 0.65148755156664\\
1.53958 0.65153977556855\\
2.89164 0.65095121035132\\
5.64454 0.65151489311025\\
13.9233 0.65080615906492\\
};
\addlegendentry{Jags};

\addplot [
color=red,
only marks,
mark=square,
mark options={solid},
forget plot
]
plot [error bars/.cd, y dir = both, y explicit]
coordinates{
(0.45066,0.65218035294639) +- (0.0,0.00234192587424542)(0.71125,0.65148755156664) +- (0.0,0.002021861014745)(1.53958,0.65153977556855) +- (0.0,0.00288806887838922)(2.89164,0.65095121035132) +- (0.0,0.00171002866112235)(5.64454,0.65151489311025) +- (0.0,0.00204304240500919)(13.9233,0.65080615906492) +- (0.0,0.00182519388205526)};
\node[above, right, inner sep=0mm, text=black]
at (axis cs:0.45066,0.65218035294639,0) {150};
\node[above, right, inner sep=0mm, text=black]
at (axis cs:0.71125,-3.34851244843336,0) {300};
\node[above, right, inner sep=0mm, text=black]
at (axis cs:1.53958,0.65153977556855,0) {750};
\node[above, right, inner sep=0mm, text=black]
at (axis cs:2.89164,0.65095121035132,0) {1500};
\node[above, right, inner sep=0mm, text=black]
at (axis cs:5.64454,0.65151489311025,0) {3000};
\node[above, right, inner sep=0mm, text=black]
at (axis cs:13.9233,0.65080615906492,0) {7500};
\end{axis}
\end{tikzpicture}%

\caption{Result of Multivariate Regression on the winequality-red data set.}
\label{figRed}
\end{center}
\vskip -0.2in
\end{figure} 

\begin{figure}[t]
\vskip 0.2in
\begin{center}

%
%
\begin{tikzpicture}[scale=0.45,font=\Large]

\begin{axis}[%
width=6.01947782842934in,
height=4.54928278277609in,
scale only axis,
xmin=-3,
xmax=60,
xlabel={Runtime (seconds)},
ymin=-10,
ymax=175,
ylabel={RMSE},
title={RMSE v. Training Time (winequality-white)},
axis x line*=bottom,
axis y line*=left,
legend style={draw=black,fill=white,legend cell align=left}
]
\addplot [
color=red,
dashed,
line width=1.0pt,
mark=square*,
mark options={solid,fill=red,draw=black}
]
table[row sep=crcr]{
8.997 164.4251236732\\
9.03956 117.146110248509\\
9.18471 8.6085734633666\\
9.43933 3.8580135069996\\
9.94786 1.13101386678362\\
11.46636 0.90436795985795\\
};
\addlegendentry{Augur};

\addplot [
color=red,
only marks,
mark=square,
mark options={solid},
forget plot
]
plot [error bars/.cd, y dir = both, y explicit]
coordinates{
(8.997,164.4251236732) +- (0.0,3.91928908402542)(9.03956,117.146110248509) +- (0.0,6.2250376010534)(9.18471,8.6085734633666) +- (0.0,0.620306111669514)(9.43933,3.8580135069996) +- (0.0,0.254703309699603)(9.94786,1.13101386678362) +- (0.0,0.0378097238472013)(11.46636,0.90436795985795) +- (0.0,0.0117312990532314)};
\node[above, right, inner sep=0mm, text=black]
at (axis cs:8.997,164.4251236732,0) {100};
\node[above, right, inner sep=0mm, text=black]
at (axis cs:9.03956,117.146110248509,0) {200};
\node[above, right, inner sep=0mm, text=black]
at (axis cs:9.18471,8.6085734633666,0) {500};
\node[above, right, inner sep=0mm, text=black]
at (axis cs:4.43933,3.8580135069996,0) {1000};
\node[above, right, inner sep=0mm, text=black]
at (axis cs:9.94786,2.13101386678362,0) {2000};
\node[above, right, inner sep=0mm, text=black]
at (axis cs:11.46636,-7.09563204014205,0) {5000};
\addplot [
color=blue,
dashed,
line width=1.0pt,
mark=*,
mark options={solid,fill=blue,draw=black}
]
table[row sep=crcr]{
1.84698 0.76133652507623\\
2.83626 0.7614541610753\\
5.93133 0.76222540987335\\
11.01971 0.76165329995683\\
21.29409 0.76183383182872\\
53.0779 0.76198911524081\\
};
\addlegendentry{Jags};

\addplot [
color=red,
only marks,
mark=square,
mark options={solid},
forget plot
]
plot [error bars/.cd, y dir = both, y explicit]
coordinates{
(1.84698,0.76133652507623) +- (0.0,0.00112853609092611)(2.83626,0.7614541610753) +- (0.0,0.00118327768136794)(5.93133,0.76222540987335) +- (0.0,0.0010481930820555)(11.01971,0.76165329995683) +- (0.0,0.000900544662231884)(21.29409,0.76183383182872) +- (0.0,0.000767288329143079)(53.0779,0.76198911524081) +- (0.0,0.00148669542258228)};
\node[below, left, inner sep=0mm, text=black]
at (axis cs:1.84698,10.7613365250762,0) {150};
\node[below, left, inner sep=0mm, text=black]
at (axis cs:2.83626,0.7614541610753,0) {300};
\node[below, left, inner sep=0mm, text=black]
at (axis cs:5.93133,0.76222540987335,0) {750};
\node[below, left, inner sep=0mm, text=black]
at (axis cs:11.01971,0.76165329995683,0) {1500};
\node[below, left, inner sep=0mm, text=black]
at (axis cs:21.29409,0.76183383182872,0) {3000};
\node[below, left, inner sep=0mm, text=black]
at (axis cs:53.0779,0.76198911524081,0) {7500};
\end{axis}
\end{tikzpicture}%

\caption{Result of Multivariate Regression on the winequality-white data set.}
\label{figWhite}
\end{center}
\vskip -0.2in
\end{figure} 

\begin{figure}[t]
\vskip 0.2in
\begin{center}

%
%
\begin{tikzpicture}[scale=0.45,font=\Large]

\begin{axis}[%
width=6.01947782842934in,
height=4.54928278277609in,
scale only axis,
xmin=-3,
xmax=35,
xlabel={Runtime (seconds)},
ymin=8,
ymax=22,
ylabel={RMSE},
title={RMSE v. Training Time (Yacht)},
axis x line*=bottom,
axis y line*=left,
legend style={at={(0.97,0.03)},anchor=south east,draw=black,fill=white,legend cell align=left}
]
\addplot [
color=red,
dashed,
line width=1.0pt,
mark=square*,
mark options={solid,fill=red,draw=black}
]
table[row sep=crcr]{
8.98097 16.287211373173\\
8.98713 16.330716267649\\
9.04306 16.37671742732\\
9.1174 16.385645218176\\
9.2897 16.373039761659\\
9.77447 16.293301817325\\
};
\addlegendentry{Augur};

\addplot [
color=red,
only marks,
mark=square,
mark options={solid},
forget plot
]
plot [error bars/.cd, y dir = both, y explicit]
coordinates{
(8.98097,16.287211373173) +- (0.0,0.0283899405088412)(8.98713,16.330716267649) +- (0.0,0.0385790225369433)(9.04306,16.37671742732) +- (0.0,0.0668355196936048)(9.1174,16.385645218176) +- (0.0,0.0401835861140611)(9.2897,16.373039761659) +- (0.0,0.0328612241302686)(9.77447,16.293301817325) +- (0.0,0.0764069840763355)};
\node[below, right, inner sep=0mm, text=black]
at (axis cs:6.98097,16.287211373173,0) {100};
\node[below, right, inner sep=0mm, text=black]
at (axis cs:8.98713,16.330716267649,0) {};
\node[below, right, inner sep=0mm, text=black]
at (axis cs:9.04306,16.37671742732,0) {};
\node[below, right, inner sep=0mm, text=black]
at (axis cs:9.1174,16.385645218176,0) {};
\node[below, right, inner sep=0mm, text=black]
at (axis cs:9.2897,16.373039761659,0) {};
\node[below, right, inner sep=0mm, text=black]
at (axis cs:9.77447,16.293301817325,0) {5000};
\addplot [
color=blue,
dashed,
line width=1.0pt,
mark=*,
mark options={solid,fill=blue,draw=black}
]
table[row sep=crcr]{
0.05799 9.687127256811\\
0.09174 9.6850729183089\\
0.19558 9.6866087237417\\
0.36768 9.715542674471\\
0.70642 9.6673206244013\\
1.75003 9.7005205490677\\
};
\addlegendentry{Jags};

\addplot [
color=red,
only marks,
mark=square,
mark options={solid},
forget plot
]
plot [error bars/.cd, y dir = both, y explicit]
coordinates{
(0.05799,9.687127256811) +- (0.0,0.0661542006722549)(0.09174,9.6850729183089) +- (0.0,0.076934514229535)(0.19558,9.6866087237417) +- (0.0,0.0926875200580533)(0.36768,9.715542674471) +- (0.0,0.122040478631181)(0.70642,9.6673206244013) +- (0.0,0.072497693860587)(1.75003,9.7005205490677) +- (0.0,0.0984720004732687)};
\node[above, right, inner sep=0mm, text=black]
at (axis cs:-1.94201,9.687127256811,0) {150};
\node[above, right, inner sep=0mm, text=black]
at (axis cs:0.09174,9.6850729183089,0) {};
\node[above, right, inner sep=0mm, text=black]
at (axis cs:0.19558,9.6866087237417,0) {};
\node[above, right, inner sep=0mm, text=black]
at (axis cs:0.36768,9.715542674471,0) {};
\node[above, right, inner sep=0mm, text=black]
at (axis cs:0.70642,9.6673206244013,0) {};
\node[above, right, inner sep=0mm, text=black]
at (axis cs:1.75003,9.7005205490677,0) {7500};
\addplot [
color=black,
dashed,
line width=1.0pt,
mark=diamond*,
mark options={solid,fill=black,draw=black}
]
table[row sep=crcr]{
21.3203 20.361825046192\\
21.38563 20.329975832094\\
22.16 20.350454103449\\
23.24503 20.355252545153\\
25.25288 20.34765770852\\
30.85102 20.346421599219\\
};
\addlegendentry{Stan};

\addplot [
color=red,
only marks,
mark=square,
mark options={solid},
forget plot
]
plot [error bars/.cd, y dir = both, y explicit]
coordinates{
(21.3203,20.361825046192) +- (0.0,0.0261307275812754)(21.38563,20.329975832094) +- (0.0,0.0383020582989069)(22.16,20.350454103449) +- (0.0,0.0316802267952099)(23.24503,20.355252545153) +- (0.0,0.0210908705724648)(25.25288,20.34765770852) +- (0.0,0.0181049724974756)(30.85102,20.346421599219) +- (0.0,0.01703589327904)};
\node[above, right, inner sep=0mm, text=black]
at (axis cs:19.3203,20.361825046192,0) {100};
\node[above, right, inner sep=0mm, text=black]
at (axis cs:21.38563,20.329975832094,0) {};
\node[above, right, inner sep=0mm, text=black]
at (axis cs:22.16,20.350454103449,0) {};
\node[above, right, inner sep=0mm, text=black]
at (axis cs:23.24503,20.355252545153,0) {};
\node[above, right, inner sep=0mm, text=black]
at (axis cs:25.25288,20.34765770852,0) {};
\node[above, right, inner sep=0mm, text=black]
at (axis cs:30.85102,20.346421599219,0) {5000};
\end{axis}
\end{tikzpicture}%

\caption{Result of Multivariate Regression on the Yacht Hydrodynamics data set.}
\label{figYacht}
\end{center}
\vskip -0.2in
\end{figure} 

The RMSEs of JAGS and Augur converge to approximately similar values, though Augur takes longer to converge
(in terms of the number of samples, and total runtime) as Metropolis-Hastings is a less efficient inference
algorithm for regression than a tuned Gibbs sampler. As mentioned
in section 5 of the paper JAGS has a special case for working with linear regression models which alters the 
sampling procedure, and this feature is not currently available in Augur.

We find that the regression results show that Augur is competitive with other systems, though linear regression
models tend not to be large enough to properly exploit all the computation available in the GPU.

\subsection{Gaussian Mixture Model}

The Gaussian Mixture Model results presented in section 5 of the paper show how each of the three systems
scale as the dataset size is increased. We sampled $100,000$ datapoints from two different mixture distributions:
one with 4 gaussians centered at \{-5,-1,1,5\} with standard deviation \{1,0.1,2,1\}, and one with 3 gaussians
centered at \{-5, 0, 5\} with standard deviations \{0.1,0.1,0.1\}. Each dataset had a flat mixing distribution, 
that is draws from each gaussian were equiprobable. From each dataset we subsampled smaller datasets using 100,
1000 and $10,000$ datapoints. 

We used the GMM presented in the paper for Augur, for Stan we used the GMM listed in the modelling handbook,
and for JAGS we wrote a standard GMM (shown in figure \ref{figJAGSGMM}, based upon Augur's. Each model used the same prior distributions and
hyperparameters.

\begin{figure}
\vskip 0.2in
\begin{center}
\begin{minipage}{0.45\textwidth}
\begin{lstlisting}
model {
    for (i in 1:N){
        z[i] ~ dcat(theta)
        y[i] ~ dnorm(mu[z[i]],sigma[z[i]])
    }
    theta[1:K] ~ ddirch(alpha)
    for (k in 1:K) {
        alpha[k] <- 1
        mu[k] ~ dnorm(0,0.01)
        sigma[k] ~ dgamma(1,1)
    }
}
\end{lstlisting}
\end{minipage}
\caption{GMM in Jags}
\label{figJAGSGMM}
\end{center}
\vskip -0.2in
\end{figure}

\begin{figure}[t]
\vskip 0.2in
\begin{center}

%
%
\begin{tikzpicture}[scale=0.45,font=\Large]

\begin{axis}[%
width=6.01947782842934in,
height=4.54928278277609in,
unbounded coords=jump,
scale only axis,
xmin=2,
xmax=5,
xtick={2,2.5,3,3.5,4,4.5,5},
xticklabels={$10^2$,,$10^3$,,$10^4$,,$10^5$},
xlabel={Dataset Size (\#points)},
ymin=0,
ymax=20,
ylabel={Runtime (minutes)},
title={Scalability (Data Set One)},
axis x line*=bottom,
axis y line*=left,
legend style={at={(0.03,0.97)},anchor=north west,draw=black,fill=white,legend cell align=left}
]
\addplot [
color=red,
dashed,
line width=1.0pt,
mark=square*,
mark options={solid,fill=red,draw=black}
]
table[row sep=crcr]{
2 0.156416666666667\\
3 0.1626\\
4 0.3448\\
5 4.21683333333333\\
};
\addlegendentry{Augur};

\addplot [
color=blue,
dashed,
line width=1.0pt,
mark=*,
mark options={solid,fill=blue,draw=black}
]
table[row sep=crcr]{
2 0.0051\\
3 0.0601166666666667\\
4 0.932416666666667\\
5 16.0780666666667\\
};
\addlegendentry{Jags};

\addplot [
color=black,
dashed,
line width=1.0pt,
mark=diamond*,
mark options={solid,fill=black,draw=black}
]
table[row sep=crcr]{
2 0.369583333333333\\
3 2.9705\\
4 14.0511833333333\\
5 inf\\
};
\addlegendentry{Stan};

\end{axis}
\end{tikzpicture}%

\caption{Evolution of runtime to draw a thousand samples for varying
  data set sizes for a Gaussian Mixture Model. Stan's $100,000$ data point was not generated.}
\label{figGMMResult}
\end{center}
\vskip -0.2in 
\end{figure}
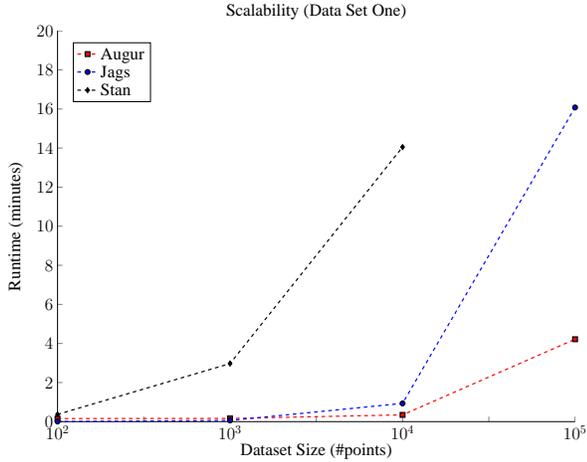

Figure 4 in the paper is from the dataset with 4 centres. In figure \ref{figGMMResult}
we show the runtime of the remaining dataset with 3 centres. For computational reasons we stopped Stan's final 
run after 3 hours (Stan took approx. 6 hours to complete on the first dataset). Here we can see that
Augur's runtime scales much more slowly as the dataset size is increased. JAGS remains reasonably competitive
until $100,000$ data points, at which point Augur is faster by a factor of 7. Stan is also relatively competitive
but scales extremely poorly as the number of datapoints is increased.

\subsection{LDA}
\begin{figure}
\vskip 0.2in
\begin{center}

%
%
\begin{tikzpicture}[scale=0.45,font=\Large]

\begin{axis}[%
width=5.96122481718648in,
height=4.54928278277609in,
scale only axis,
xmin=0,
xmax=4.5,
xtick={0,1,2,3,4},
xticklabels={1,10,$10^2$,$10^3$,$10^4$},
xlabel={Runtime (seconds)},
ymin=-165000,
ymax=-125000,
ylabel={$\log_{10}$ Predictive Probability},
title={Predictive Probability v. Training Time},
axis x line*=bottom,
axis y line*=left,
legend style={draw=black,fill=white,legend cell align=left}
]
\addplot [
color=red,
dashed,
line width=1.0pt,
mark size=1.8pt,
mark=square*,
mark options={solid,fill=red,draw=black}
]
table[row sep=crcr]{
0.926342446625655 -142385.183069909\\
0.948901760970214 -142318.736014178\\
0.9804578922761 -142420.795217425\\
1.04688519083771 -142890.701846844\\
1.14921911265538 -143146.501296685\\
1.30319605742049 -141502.696682681\\
1.5081255360832 -138379.250768833\\
1.7523558041535 -134792.412642794\\
2.02135471308142 -131991.213234518\\
2.30569522891186 -130390.403774223\\
};
\addlegendentry{Augur};

\node[above, inner sep=0mm, text=black]
at (axis cs:0.826342446625655,-145385.183069909,0) {1};
\node[above, inner sep=0mm, text=black]
at (axis cs:0.898901760970214,-145318.736014178,0) {2};
\node[above, inner sep=0mm, text=black]
at (axis cs:0.9804578922761,-145420.795217425,0) {$\text{2}^\text{2}$};
\node[above, inner sep=0mm, text=black]
at (axis cs:1.04688519083771,-145890.701846844,0) {$\text{2}^\text{3}$};
\node[above, inner sep=0mm, text=black]
at (axis cs:1.14921911265538,-146146.501296685,0) {$\text{2}^\text{4}$};
\node[above, inner sep=0mm, text=black]
at (axis cs:1.30319605742049,-144502.696682681,0) {$\text{2}^\text{5}$};
\node[above, inner sep=0mm, text=black]
at (axis cs:1.5081255360832,-141379.250768833,0) {$\text{2}^\text{6}$};
\node[above, inner sep=0mm, text=black]
at (axis cs:1.7523558041535,-137792.412642794,0) {$\text{2}^\text{7}$};
\node[above, inner sep=0mm, text=black]
at (axis cs:2.02135471308142,-134991.213234518,0) {$\text{2}^\text{8}$};
\node[above, inner sep=0mm, text=black]
at (axis cs:2.30569522891186,-133390.403774223,0) {$\text{2}^\text{9}$};
\addplot [
color=red,
only marks,
mark=square,
mark options={solid},
forget plot
]
plot [error bars/.cd, y dir = both, y explicit]
coordinates{
(0.926342446625655,-142385.183069909) +- (0.0,826.462399061886)(0.948901760970214,-142318.736014178) +- (0.0,885.526448600722)(0.9804578922761,-142420.795217425) +- (0.0,880.314914817893)(1.04688519083771,-142890.701846844) +- (0.0,797.798963256268)(1.14921911265538,-143146.501296685) +- (0.0,743.946447500261)(1.30319605742049,-141502.696682681) +- (0.0,731.351907525066)(1.5081255360832,-138379.250768833) +- (0.0,679.236569696688)(1.7523558041535,-134792.412642794) +- (0.0,821.685159760964)(2.02135471308142,-131991.213234518) +- (0.0,811.696386677184)(2.30569522891186,-130390.403774223) +- (0.0,708.334299984199)};
\addplot [
color=blue,
dashed,
line width=1.0pt,
mark size=2.5pt,
mark=*,
mark options={solid,fill=blue,draw=black}
]
table[row sep=crcr]{
0.227886704613674 -139747.71268131\\
0.313867220369153 -140255.402930655\\
0.450249108319361 -140922.479254859\\
0.636487896353365 -141247.331527322\\
0.864511081058392 -139363.362064826\\
1.12385164096709 -136699.834007313\\
1.40174508223706 -134613.049021768\\
1.68957521575994 -132916.260480972\\
1.98349097181517 -131421.853168743\\
2.28053284445138 -130658.363469557\\
};
\addlegendentry{Cuda};

\node[above, inner sep=0mm, text=black]
at (axis cs:0.227886704613674,-138747.71268131,0) {1};
\node[above, inner sep=0mm, text=black]
at (axis cs:0.313867220369153,-139255.402930655,0) {2};
\node[above, inner sep=0mm, text=black]
at (axis cs:0.450249108319361,-139922.479254859,0) {$\text{2}^\text{2}$};
\node[above, inner sep=0mm, text=black]
at (axis cs:0.636487896353365,-140247.331527322,0) {$\text{2}^\text{3}$};
\node[above, inner sep=0mm, text=black]
at (axis cs:0.864511081058392,-138363.362064826,0) {$\text{2}^\text{4}$};
\node[above, inner sep=0mm, text=black]
at (axis cs:1.12385164096709,-135699.834007313,0) {$\text{2}^\text{5}$};
\node[above, inner sep=0mm, text=black]
at (axis cs:1.40174508223706,-133613.049021768,0) {$\text{2}^\text{6}$};
\node[above, inner sep=0mm, text=black]
at (axis cs:1.68957521575994,-131916.260480972,0) {$\text{2}^\text{7}$};
\node[above, inner sep=0mm, text=black]
at (axis cs:1.98349097181517,-130421.853168743,0) {$\text{2}^\text{8}$};
\node[above, inner sep=0mm, text=black]
at (axis cs:2.28053284445138,-129658.363469557,0) {$\text{2}^\text{9}$};
\addplot [
color=blue,
only marks,
mark=o,
mark options={solid},
forget plot
]
plot [error bars/.cd, y dir = both, y explicit]
coordinates{
(0.227886704613674,-139747.71268131) +- (0.0,861.640252096055)(0.313867220369153,-140255.402930655) +- (0.0,826.028104579978)(0.450249108319361,-140922.479254859) +- (0.0,852.085773494182)(0.636487896353365,-141247.331527322) +- (0.0,800.839024629589)(0.864511081058392,-139363.362064826) +- (0.0,907.241172695882)(1.12385164096709,-136699.834007313) +- (0.0,874.234792071249)(1.40174508223706,-134613.049021768) +- (0.0,789.113073618209)(1.68957521575994,-132916.260480972) +- (0.0,693.568287599483)(1.98349097181517,-131421.853168743) +- (0.0,681.408042106195)(2.28053284445138,-130658.363469557) +- (0.0,729.180435115559)};
\addplot [
color=black,
dashed,
line width=1.0pt,
mark size=4.3pt,
mark=diamond*,
mark options={solid,fill=red,draw=black}
]
table[row sep=crcr]{
1.35391623092036 -159122.89240246\\
1.60573589387675 -161205.334443186\\
1.86153441085904 -161942.766473458\\
2.11304041808782 -159487.699767259\\
2.37540753330879 -154857.686295688\\
2.64616837877408 -150875.640191117\\
2.92670249418265 -148885.268580555\\
3.21581991297026 -147799.098081315\\
3.50458366925724 -147437.330777889\\
3.80279629434003 -147155.039364652\\
};
\addlegendentry{Factorie(Collapsed)};

\node[above, inner sep=0mm, text=black]
at (axis cs:1.35391623092036,-158122.89240246,0) {1};
\node[above, inner sep=0mm, text=black]
at (axis cs:1.60573589387675,-160205.334443186,0) {2};
\node[above, inner sep=0mm, text=black]
at (axis cs:1.86153441085904,-160942.766473458,0) {$\text{2}^\text{2}$};
\node[above, inner sep=0mm, text=black]
at (axis cs:2.11304041808782,-158487.699767259,0) {$\text{2}^\text{3}$};
\node[above, inner sep=0mm, text=black]
at (axis cs:2.37540753330879,-153857.686295688,0) {$\text{2}^\text{4}$};
\node[above, inner sep=0mm, text=black]
at (axis cs:2.64616837877408,-149875.640191117,0) {$\text{2}^\text{5}$};
\node[above, inner sep=0mm, text=black]
at (axis cs:2.92670249418265,-147885.268580555,0) {$\text{2}^\text{6}$};
\node[above, inner sep=0mm, text=black]
at (axis cs:3.21581991297026,-146799.098081315,0) {$\text{2}^\text{7}$};
\node[above, inner sep=0mm, text=black]
at (axis cs:3.50458366925724,-146437.330777889,0) {$\text{2}^\text{8}$};
\node[above, inner sep=0mm, text=black]
at (axis cs:3.80279629434003,-146155.039364652,0) {$\text{2}^\text{9}$};
\addplot [
color=black,
only marks,
mark=diamond,
mark options={solid},
forget plot
]
plot [error bars/.cd, y dir = both, y explicit]
coordinates{
(1.35391623092036,-159122.89240246) +- (0.0,1039.70098967638)(1.60573589387675,-161205.334443186) +- (0.0,1011.90614283457)(1.86153441085904,-161942.766473458) +- (0.0,935.036019537714)(2.11304041808782,-159487.699767259) +- (0.0,1029.27792211072)(2.37540753330879,-154857.686295688) +- (0.0,923.744363008213)(2.64616837877408,-150875.640191117) +- (0.0,846.005650747538)(2.92670249418265,-148885.268580555) +- (0.0,902.46393339496)(3.21581991297026,-147799.098081315) +- (0.0,823.422337688564)(3.50458366925724,-147437.330777889) +- (0.0,791.718840509624)(3.80279629434003,-147155.039364652) +- (0.0,858.600190722733)};
\end{axis}
\end{tikzpicture}%

\caption{Average over 10 runs of the evolution of the predictive
  probability over time.}
\label{Varying}
\end{center}
\vskip -0.2in
\end{figure}
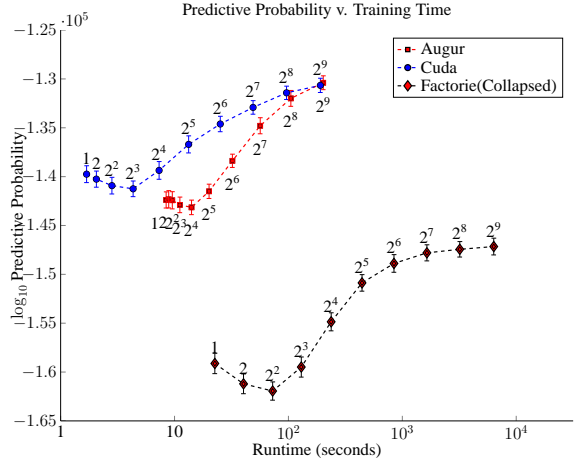

In an attempt to confirm the result presented in the paper, we present
another result (figure \ref{Varying}) measuring the predictive probability
averaging across multiple runs using different train/test splits.
In this experiment, we averaged across 10 runs with different
train/test splits and present the timings with error bars.  We also
ran an experimment across 10 different random initializations and seeds, and all algorithms
again showed robustness to the variation. We reduced the
maximum number of samples to 512 as generating results for the
Collapsed Gibbs sampler was proving prohibitive in terms of runtime
for repeated experiments.

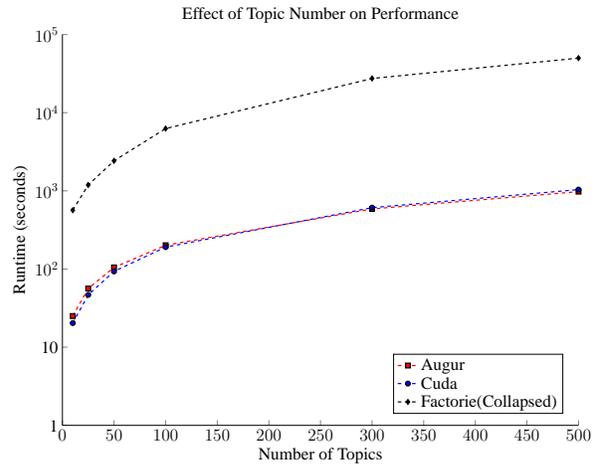
\begin{figure}
\vskip 0.2in
\begin{center}

%
%
\begin{tikzpicture}[scale=0.45,font=\Large]

\begin{axis}[%
width=6.00283411093138in,
height=4.54928278277609in,
scale only axis,
xmin=0,
xmax=500,
xlabel={Number of Topics},
ymin=0,
ymax=5,
ytick={0,1,2,3,4,5},
yticklabels={1,10,$10^2$,$10^3$,$10^4$,$10^5$},
ylabel={Runtime (seconds)},
title={Effect of Topic Number on Performance},
axis x line*=bottom,
axis y line*=left,
legend style={at={(0.97,0.03)},anchor=south east,draw=black,fill=white,legend cell align=left}
]
\addplot [
color=red,
dashed,
line width=1.0pt,
mark=square*,
mark options={solid,fill=red,draw=black}
]
table[row sep=crcr]{
10 1.3981136917305\\
25 1.75027691515399\\
50 2.02069267868203\\
100 2.30414571276379\\
300 2.76866769441187\\
500 2.98996568084046\\
};
\addlegendentry{Augur};

\addplot [
color=blue,
dashed,
line width=1.0pt,
mark=*,
mark options={solid,fill=blue,draw=black}
]
table[row sep=crcr]{
10 1.30835094858673\\
25 1.66940986728778\\
50 1.96946250797464\\
100 2.28060113153691\\
300 2.78502358504569\\
500 3.01597136117771\\
};
\addlegendentry{Cuda};

\addplot [
color=black,
dashed,
line width=1.0pt,
mark=diamond*,
mark options={solid,fill=black,draw=black}
]
table[row sep=crcr]{
10 2.75272434485752\\
25 3.07494437051221\\
50 3.38430322450415\\
100 3.79680390746333\\
300 4.43759853314498\\
500 4.69840688061967\\
};
\addlegendentry{Factorie(Collapsed)};

\end{axis}
\end{tikzpicture}%

\caption{Comparison of scalability of Augur, hand-written CUDA, and
  Factorie's collapsed Gibbs {\it w.r.t} the number of topics.}
\label{Topics}
\end{center}
\vskip -0.2in
\end{figure}

A third experiment (figure \ref{Topics}) reports on the natural
logarithm of run time in milliseconds to draw 512 samples as the
number of topics varies. The sparse implementation's running time does
not increase as quickly as Augur's as the number of topics
increases. As a result, it runs faster when the number of topics is
large. This is because Augur's Gibbs sampler is linear in the number
of topics during the step of sampling each of the $z_{ij}$.  The
collapsed Gibbs sampler's performance worsen when the number of topics
is increased, as seen in our results and in the experiments in
\cite{OnlineLDA}. Again, Augur's generated code is on par with the
hand-written CUDA implementation.

We experimented with the SparseLDA implementation which forms Factorie's standard
LDA model, but this implementation proved to be unreliable. The predictive probability
measure actually decreased as more samples were drawn using the SparseLDA implementation.
We are working with the developers of Factorie to investigate this problem. The SparseLDA
implementation is interesting as it uses a set of LDA specific assumptions to generate a
highly optimised Gibbs sampler. We found Augur to be competitive in terms of runtime when drawing
more than 256 samples. With smaller sample sizes there is insufficient computation to amortize
the compilation costs.

\end{document}